\documentclass[letterpaper]{article} 
\usepackage{aaai24}  
\usepackage{times}  
\usepackage{helvet}  
\usepackage{courier}  
\usepackage[hyphens]{url}  
\usepackage{graphicx} 
\usepackage{natbib}  
\usepackage{caption} 
\usepackage{booktabs}
\usepackage{algorithm}
\usepackage{algorithmic}

\usepackage{newfloat}
\usepackage{listings}
\DeclareCaptionStyle{ruled}{labelfont=normalfont,labelsep=colon,strut=off} 
\floatstyle{ruled}
\newfloat{listing}{tb}{lst}{}
\floatname{listing}{Listing}
\usepackage{cite}
\usepackage{color,xcolor}
\usepackage{epsfig}
\usepackage{graphicx}
\usepackage{adjustbox}
\usepackage{array}
\usepackage{booktabs}
\usepackage{colortbl}
\usepackage{float}
\usepackage{hhline}
\usepackage{subcaption} %
\usepackage[font=footnotesize]{caption}
\usepackage[toc,page]{appendix}
\usepackage{multirow}

\usepackage{amsmath,amsfonts,amsthm,amssymb}
\usepackage{mathtools}
\usepackage{bm}
\usepackage{nicefrac}
\usepackage{microtype}

\usepackage{changepage}
\usepackage{extramarks}
\usepackage{fancyhdr}
\usepackage{lastpage}
\usepackage{soul}
\usepackage{xspace}
\usepackage{indentfirst}
\usepackage{pifont}
\usepackage{cuted}

\usepackage{url}

\usepackage{algorithm, algorithmic}
\usepackage{enumitem}

\usepackage{wasysym}
\allowdisplaybreaks[4]

\newcolumntype{L}[1]{>{\raggedright\let\newline\\\arraybackslash\hspace{0pt}}m{#1}}
\newcolumntype{C}[1]{>{\centering\let\newline\\\arraybackslash\hspace{0pt}}m{#1}}
\newcolumntype{R}[1]{>{\raggedleft\let\newline\\\arraybackslash\hspace{0pt}}m{#1}}

\newcommand{\sect}[1]{Section~\ref{sect:#1}}
\newcommand{\app}[1]{Appendix~\ref{app:#1}}

\newcommand{\lblsect}[1]{\label{sect:#1}}
\newcommand{\lblapp}[1]{\label{app:#1}}

\newcommand{\ignorethis}[1]{}

\makeatletter
\DeclareRobustCommand\onedot{\futurelet\@let@token\@onedot}
\def\@onedot{\ifx\@let@token.\else.\null\fi\xspace}

\makeatother

\definecolor{citecolor}{RGB}{34,139,34}
\definecolor{mydarkblue}{rgb}{0,0.08,1}
\definecolor{mydarkgreen}{rgb}{0.02,0.6,0.02}
\definecolor{mydarkred}{rgb}{0.8,0.02,0.02}
\definecolor{mydarkorange}{rgb}{0.40,0.2,0.02}
\definecolor{mypurple}{RGB}{111,0,255}
\definecolor{myred}{rgb}{1.0,0.0,0.0}
\definecolor{mygold}{rgb}{0.75,0.6,0.12}
\definecolor{mydarkgray}{rgb}{0.66, 0.66, 0.66}

\title{Reducing Spatial Fitting Error in Distillation of Denoising Diffusion Models}
\author{
    Shengzhe Zhou\textsuperscript{\rm 1},
    Zejian Li\textsuperscript{\rm 2}\thanks{Corresponding author.},
    Shengyuan Zhang\textsuperscript{\rm 3},
    Lefan Hou\textsuperscript{\rm 4},
    Changyuan Yang\textsuperscript{\rm 5},
    Guang Yang\textsuperscript{\rm 6},
    Zhiyuan Yang\textsuperscript{\rm 7},
    Lingyun Sun\textsuperscript{\rm 8}
}
\affiliations{
    \textsuperscript{\rm 1, 2}School of Software Technology, Zhejiang University\\
    \textsuperscript{\rm 3, 4, 8} College of Computer Science and Technology, Zhejiang University\\

     \textsuperscript{\rm 5, 6, 7} Alibaba Group\\
\{zhoujj7248, zejianlee, zhangshengyuan, houlefan, sunly\}@zju.edu.cn, \{changyuan.yangcy, adam.yzy\}@alibaba-inc.com, qingyun@taobao.com
}

\begin{document}

\maketitle

\begin{abstract}
    Denoising Diffusion models have exhibited remarkable capabilities in image generation. However, generating high-quality samples requires a large number of iterations. Knowledge distillation for diffusion models is an effective method to address this limitation with a shortened sampling process but causes degraded generative quality. Based on our analysis with bias-variance decomposition and experimental observations, we attribute the degradation to the spatial fitting error occurring in the training of both the teacher and student model. Accordingly, we propose $\textbf{S}$patial $\textbf{F}$itting-$\textbf{E}$rror $\textbf{R}$eduction $\textbf{D}$istillation model ($\textbf{SFERD}$). SFERD utilizes attention guidance from the teacher model and a designed semantic gradient predictor to reduce the student's fitting error. Empirically, our proposed model facilitates high-quality sample generation in a few function evaluations. We achieve an FID of 5.31 on CIFAR-10 and 9.39 on ImageNet 64$\times$64 with only one step, outperforming existing diffusion methods. Our study provides a new perspective on diffusion distillation by highlighting the intrinsic denoising ability of models. Project link: \url{https://github.com/Sainzerjj/SFERD}.
\end{abstract}

\section{Introduction}

Diffusion-based (DPMs)~\cite{ho2020denoising, sohl2015deep, song2019generative} and score-based~\cite{song2020score} generative models have recently achieved outstanding performance in synthesizing high-quality images. They have already shown comparable or even superior results to GAN~\cite{goodfellow2020generative} in multiple fields such as 3D generation~\cite{luo2021diffusion, zhou20213d}, text-to-image generation~\cite{rombach2022high, ruiz2023dreambooth}, image restoration~\cite{lugmayr2022repaint, wang2022imagen}, controllable image editing~\cite{kawar2022imagic, couairon2022diffedit} and graph generation~\cite{Huang2023ConditionalDB}.

A major problem with diffusion models is the slow sampling speed, as it often requires hundreds of iterations to achieve satisfactory generative quality. To address this issue, there are two mainstream improvements: fast sampling schemes without extra training~\cite{song2020denoising, lu2022dpm, liu2022pseudo, kong2021fast} and trained acceleration schemes~\cite{salimans2022progressive, dockhorn2022genie, zhang2022unsupervised}. The former aims to reduce errors resulting from large-step sampling by utilizing better numerical methods for integration, whereas the latter enhances performance through additional model training and fine-tuning. With the advent of large-scale pre-training models, the latter approach has become more notable. Among trained schemes, diffusion models based on knowledge distillation have demonstrated considerable potential for fast sampling. 
An illustrative example is Progressive Distillation model (PD)~\cite{salimans2022progressive}, where the student model learns the two-step inference output of the teacher model in a single step without compressing the model size. However, these diffusion models based on distillation face degraded generation quality given limited distillation sampling step.


In this paper, we investigate scalable
enhancements for the diffusion distillation model to 
improve the quality of the images generated within few steps or even a single step. 
We begin by reframing the process of diffusion distillation models and designing a multi-step training framework for distillation.
We then conduct analysis with bias-variance decomposition and preliminary experiments to identify the fitting errors of both the teacher and student models.
Our results show that fitting errors have a broad impact on model performance within our framework. 
Especially, we observe a positive correlation between the self-attention maps of the diffusion model and the spatial fitting error in the predicted noise.

\begin{figure}[t]
    \centering
    \includegraphics[width = 1.01\linewidth]{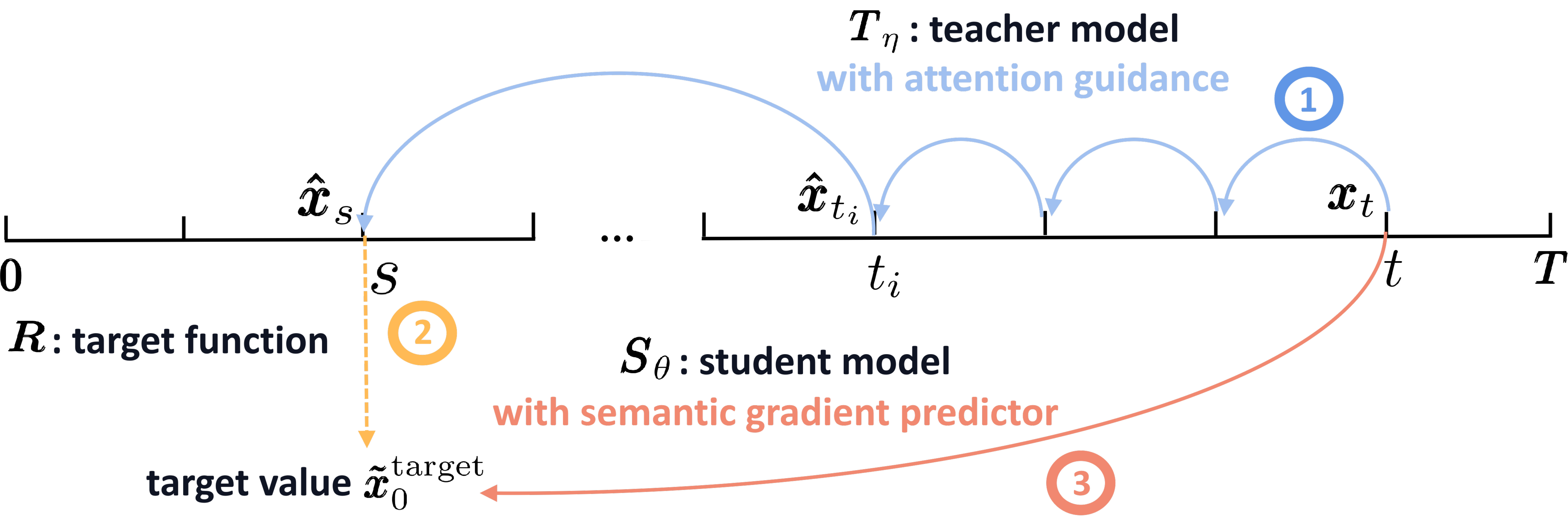}
   \caption{The process of Spatial Fitting-Error Reduction Distillation model. In the first step, $\bm{\hat x}_{s}$ is predicted by the teacher model $\bm T_\eta$ from $\bm x_t$ with \textbf{attention guidance} (\sect{DAG}). The target function $\bm {R}$ further calculates the target value $\bm x_0^{\mathrm{target}}$ based on $\bm {\hat x_s}$ in the second step. The student model $\bm S_\theta$ tries to regress the target value with \textbf{semantic gradient predictor} (\sect{DSE}) in the final step.}
    \label{distillation}
\end{figure}

Based on our observations, we propose \textbf{S}patial \textbf{F}itting-\textbf{E}rror \textbf{R}eduction \textbf{D}istillation model (\textbf{SFERD}). 
SFERD utilizes internal and external representations to reduce the fitting error of the teacher and the student models, respectively. 
First, we design attention guidance to improve the denoised prediction using intrinsic information in the self-attention maps of the teacher model.
We define the spatial regions having high self-attention scores as ``Risky Regions''.
According to the observed correlation, these regions exhibits the high fitting error of the teacher model, which is further inherited by the student model.
Therefore, by reducing the error in Risky Regions, we expect to enhance the quality of denoised prediction in the student model. 
This method does not require additional supervised information or extra auxiliary classifiers,  
and it can be combined with various diffusion models as well. 
Inspired by the classifier-based gradient guidance~\cite{dhariwal2021diffusion}, we also introduce a semantic gradient predictor and reformulate the training loss for the student model. The design is to reduce the student model's fitting error by providing additional information for image reconstruction from a learned latent space. 

Empirically, SFERD efficiently reduces the fitting error of the student model, leading to superior performance as compared to other distillation models on CIFAR-10~\cite{Krizhevsky2009LearningML} and ImageNet 64$\times$64~\cite{Deng2009ImageNetAL}. Notably, it achieves single-step FID scores of 9.39 and 5.31 for ImageNet 64$\times$64 and CIFAR-10 respectively.
Furthermore, finetuning pre-trained diffusion model itself with our proposed method also results in improved performance. 
 
\section{Preliminary}
We briefly review backgrounds of diffusion distillation. 
Detailed content is deferred to \app{related-work}. 
Diffusion models typically include the forward process and the backward denoising process. Ho et al.~\cite{ho2020denoising} define the forward diffusion process as $ q(\bm{x}_{t} | \bm{x}_{t-1}) = \mathcal{N} (\bm{x}_{t}; \sqrt{1-\beta_{t}}\bm{x}_{t-1} \,,\, \beta_{t}\bm{\mathrm{I}})$, where $\{\beta_{t}\}_{t=1}^T$ is a variance schedule used to control the noise intensity. Since the process satisfies Markov conditions, the forward process can be expressed as $q(\bm{x}_{1:T} | \bm{x}_{0}) = \prod_{t=1}^{T} q(\bm{x}_{t} |\bm {x}_{t-1}) \, $.
\begin{equation}
\label{objective}
\begin{aligned}
q(\bm{x}_{t} | \bm{x}_{0}) = \mathcal{N} (\bm{x}_{t}; \sqrt{\bar{\alpha}_{t}}\bm{x}_{0} \,,\, (1-\bar{\alpha}_{t})\bm{\mathrm{I}})
\end{aligned}
\end{equation}
where $\bar{\alpha}_{t}=1-\beta_{t}\,,\,\bar{\alpha}_{t}=\prod_{i=1}^{t}\alpha_{i}$. The sampling procedure becomes $p_{\theta}(\bm{x}_{t-1} | \bm{x}_{t}) = \mathcal{N} (\bm{x}_{t-1}; \bm{\mu}_{\theta}(\bm{x}_{t},t) \, \, ,\bm{\Sigma}_{\theta}(\bm{x}_{t},t))$. Its training objective is to minimize the upper bound of the variance of the negative log-likelihood, allowing for various predicted parameters, such as $\epsilon$-prediction~\cite{song2021maximum}, v-prediction~\cite{ho2022imagen}, $x_0$-prediction~\cite{salimans2022progressive}. To ensure a uniform representation, we define the training loss in distillation as:
\begin{equation}\label{DPM_train_loss}
\begin{split}
\mathbb{E}_{t \sim U[0,1]\,,\, \bm{x}_0 \sim p(\bm{x})\,,\,\bm{x}_t \sim q(\bm{x}_t|
\bm {x}_0)}[\omega(\lambda_t) \| \bm{\hat x}_0(\bm{x}_t\,,\,t; \theta)-\bm{x}_0 \|_2^2]
\end{split}
\end{equation}
where $\bm{\hat x}_0$ denotes denoise prediction, $\lambda_t=\log[\bar \alpha_t/(1-\bar \alpha_{t})]$ represents the signal-to-noise ratio~\cite{kingma2021variational}, and different choices of the weighting function $\omega(\cdot)$ correspond to different predicted variable $\mathbf{\theta}$. 

DDIM~\cite{song2020denoising} breaks the limitations of Markov chains and allows for more stable and fast reverse sampling, the deterministic process becomes: 
\begin{equation}\label{ddim_sampler}
\begin{aligned}
\bm{x}_{t-1}  =&  \sqrt{\bar{\alpha}_{t-1}} \bigg( \frac{\bm{x}_{t} - \sqrt{1 - \bar{\alpha}_{t}} \cdot \bm{\epsilon}_{\theta}(\bm{x}_{t},t)}{\sqrt{\bar{\alpha}_{t}}} \bigg)  \\ & +
\sqrt{1 - \bar{\alpha}_{t-1}} \cdot \bm{\epsilon}_{\theta}(\bm{x}_{t},t) 
\end{aligned}
\end{equation} 
DDIM has been demonstrated to be a first-order discrete numerical solution~\cite{lu2022dpm} of ordinary differential equation (ODE): 
\begin{equation}\label{ode-3}
\begin{aligned}
\bm x_t =& \underbrace{\sqrt{\frac{\bar \alpha_t}{\bar \alpha_s}}\bm x_s -\frac{1}{2}\sqrt{\bar \alpha_t}\bm\epsilon_\theta(x_s, s)\int_s^t\bigg(\frac{d\lambda_\delta}{d\delta}\bigg)\sqrt{\frac{1-\bar \alpha_\delta}{\bar \alpha_\delta}}d\delta}_{A} \\& +
\underbrace{\mathcal{O}\big((\lambda_t-\lambda_s)^2\big)}_{B}
\end{aligned}
\end{equation}
here $\lambda_\delta=\log[\bar \alpha_\delta/(1-\bar \alpha_{\delta})]$.
If $\bm \epsilon_\theta(\bm x_s, s)$ is assumed to be constant from $s$ to $t$, the first term (A) of Eq.(\ref{ode-3}) is equivalent to Eq.(\ref{ddim_sampler})~\cite{lu2022dpm}. Such assumption brings high-order approximation error formalized in the second term (B), which leads to the discretization error in sampling process.


\begin{figure*}[t]
    \centering
        \begin{subfigure}{0.324\linewidth}
            \includegraphics[width=\linewidth]{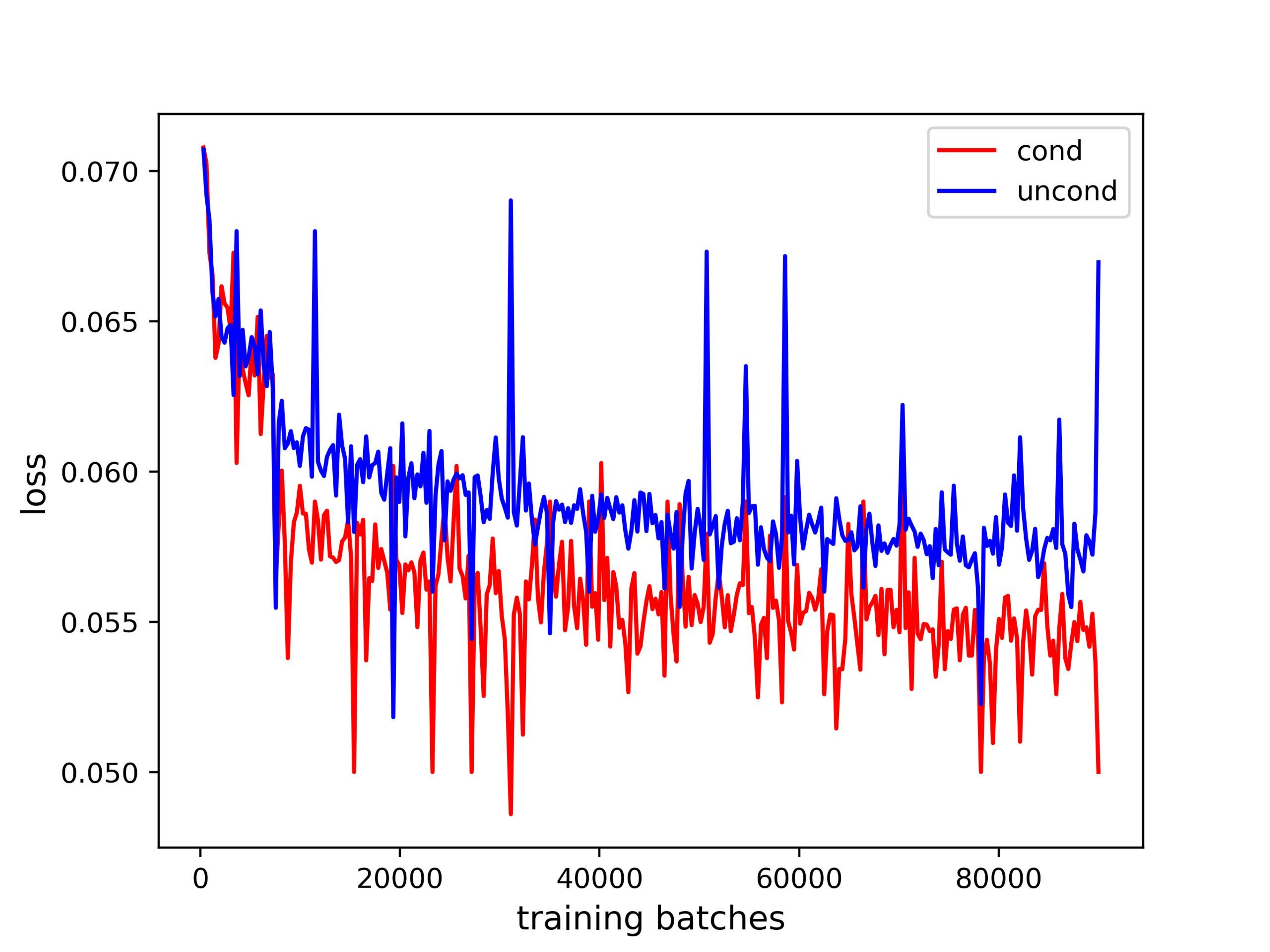}
            \caption{} 
            \label{cond_loss}
        \end{subfigure}
    \hspace{.01in}
        \begin{subfigure}{0.305\linewidth}
        
            \includegraphics[width = .945\linewidth]{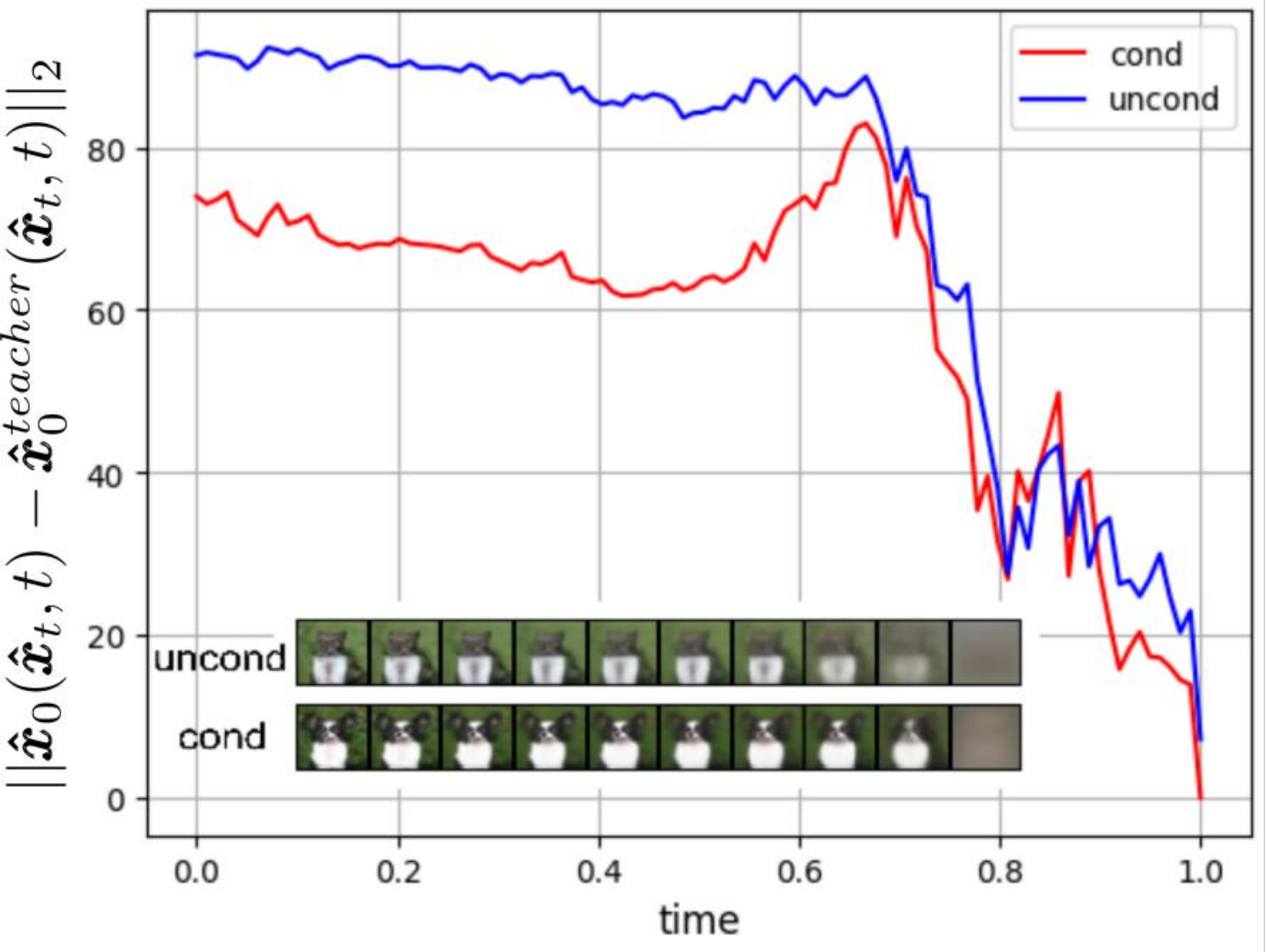}
        \caption{}
        \label{cond_delta_x0}
        \end{subfigure}
    \hspace{.01in}
        \begin{subfigure}{0.32\linewidth}
            \includegraphics[width=\linewidth]{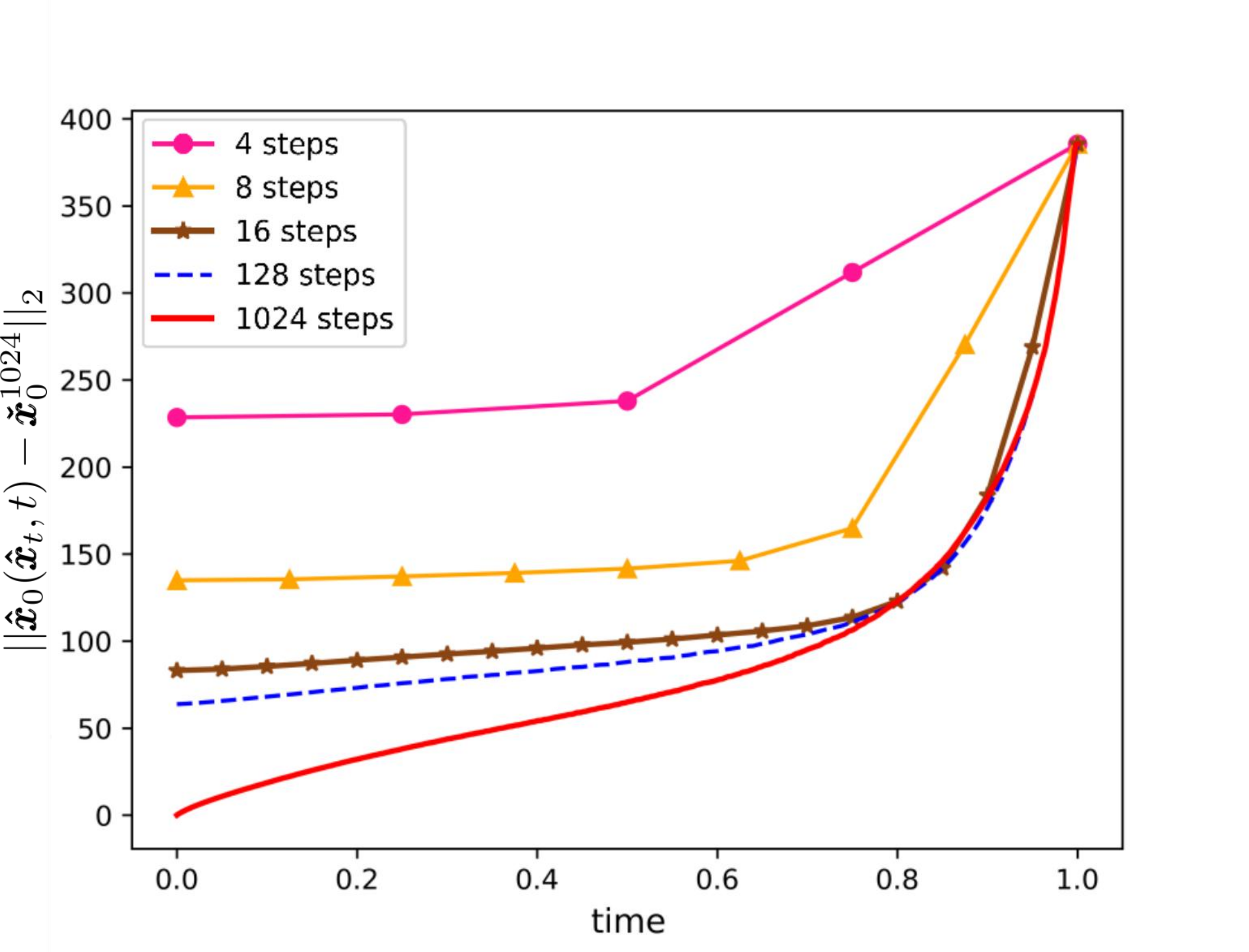}
            \caption{}
            \label{delta_x0}
        \end{subfigure}%
    \caption{(a) shows the training MSE loss of conditional and unconditional diffusion models based on $\epsilon$-prediction. (b) shows the $\ell_2$ distances of the predicted real samples $\bm{\hat x}_0^t$ between the baseline teacher model and the student models in each time step (1 $\rightarrow$ 0). Both unconditional and conditional student models are compared here.  (c) shows the $\ell_2$ distances from $\bm {\hat x}_0^t$ given by different student models in each time step (1 $\rightarrow$ 0) to the final generated sample $\check{x}_0^{1000}$ given by the baseline teacher model with the same initial noise. Examples in these figures are based on models trained on CIFAR-10~\cite{Krizhevsky2009LearningML}. These figures are best viewed magnified on the screen.}
    \label{fig:loss}
\end{figure*}

\section{Generalizing Diffusion Distillation Model}
\subsection{The Process of Diffusion Distillation Model}
\lblsect{loss-of-distillation}
The distillation of diffusion models is the process that trains the student model $\bm S_\theta$ to approximate the corresponding target distribution in a single step, bypassing the costly multi-step sampling process. Based on this comprehension, we divide the training process of the diffusion distillation model into three steps (Figure \ref{distillation}), formulated in Eq.(\ref{distillation-process}):
\begin{align} \label{distillation-process}
\hat{\bm x}_s &=\bm T_\eta^{(t-s)}(\bm x_t,s,t) \coloneqq \bm S_\theta^{(1)}(\bm x_t,s,t)
\\
\tilde{\bm x}_0^{\mathrm{target}} &=\bm R(\bm {\hat x}_{t_i},s,t_i)\,\,,t_i
\in [s\,, t-1] \label{target-function}
\end{align}

Here $\bm T_\eta$ and $\bm S_\theta$ denote the teacher model and the student model, respectively, whose superscripts represent the corresponding number of sampling steps. Note that our framework can be generalized to different diffusion models. The sampling in Eq.(\ref{distillation-process}) may use different numerical solvers $g$ for the teacher or the student, and we will introduce the choice of $g$ in implementation details.
In the first step, we sample $\bm{\hat x}_{t_i}$ with the teacher model $\bm T_\eta$ in the same ODE generation path as $\bm x_t$. 
The process ensures the consistency of the target distribution~\cite{song2023consistency}, which is crucial for effective distillation~\cite{hinton2015distilling}. 
In the second step, a target value $\bm{\tilde x}_0^{\mathrm {target}}$ is obtained for the student's learning with the target function $\bm {R}$ as Eq.~\ref{target-function}. 
The target value is typically the denoised prediction of $\bm{\hat x_s}$. $\bm {R}$ is defined differently given various distillation principles. It can be the previous teacher from the first step~\cite{salimans2022progressive, meng2022distillation, luhman2021knowledge}, another newly proposed teacher~\cite{dockhorn2022genie} or a competent student~\cite{song2023consistency}. 
$\bm T_\eta$ is a pre-trained diffusion generative model, whose network parameters $\eta$ are often fixed during the distillation process. 
Finally, in the third step, the student model $\bm S_\theta$ is trained to fit the target value $\bm {\tilde x}_0^{\mathrm {target}}$ as Eq.(\ref{training}) choosing appropriate $\omega(\lambda_t)$. 
\begin{equation}\label{training}
\begin{aligned}
\mathcal{L}_{\mathit{S}}&=\mathbb{E}_{t ,\bm{x}_0,\bm{x}_t}[\omega(\lambda_t) \| \bm {\hat x}_0^\mathrm  {student}(\bm x_t\,,\,t)-\bm{\tilde x}_0^{\mathrm {target}} \|_2^2]
\end{aligned}
\end{equation}
Such three-step division of the
distillation process provides a new insight into identifying the errors that may occur during the process and how to mitigate them. We argue that the common errors arising in the training of diffusion distillation models mainly originate from the sampling error of the teacher model in the first step, as well as the fitting error of the student model in the third step.
\subsubsection{Sampling error in the teacher model} 
The sampling error of the teacher model can be divided into the fitting error in training and the discretization error in sampling. Without loss of generality, we consider teacher model of $\bm \epsilon$-prediction, whose fitting error is reflected by the mismatch between the predicted noise $\bm \epsilon_\eta$ and the real noise $\bm \epsilon$. This error is caused by the model's limited capacity or the divergence during training. 
On the other hand, the discretization error is largely introduced in the sampling when the step size is large and the high-order variation of $\bm \epsilon_\eta$ is omitted, as in Eq.(\ref{ode-3}). 
\subsubsection{Fitting error in the student model}
The error is generally caused by the student model's inability to regress the denoised prediction $\bm{\tilde x}_0^\mathrm{target}$ when the loss in Eq.(\ref{training}) fails to converge to zero. To further investigate the main cause, we examine denoised prediction given student models with different total steps and visualize errors in Figure \ref{delta_x0}. The model with 1024 total steps is the teacher model, while others are students obtained through Progressive Distillation method. 
The curves depict the mean square error between the intermediate denoised samples and the final sample generated by the teacher. Empirical evidence shows that the error increases as the number of sampling steps decreases. The gap widens substantially when the student models sample only 4 or 8 steps. Furthermore, this error typically occurs during the middle or final stages of sampling.

Previous works have mainly reduced the discretization error of the teacher model efficiently~\cite{lu2022dpm, lu2022dpm++}. In this study, we focus on reducing the fitting errors of both the teacher and student models to pave a new way for the distillation of diffusion models.

\begin{figure*}[t]
    \centering
    \begin{minipage}[t]{0.3\linewidth}
        \centering
    \includegraphics[width=0.95\linewidth]{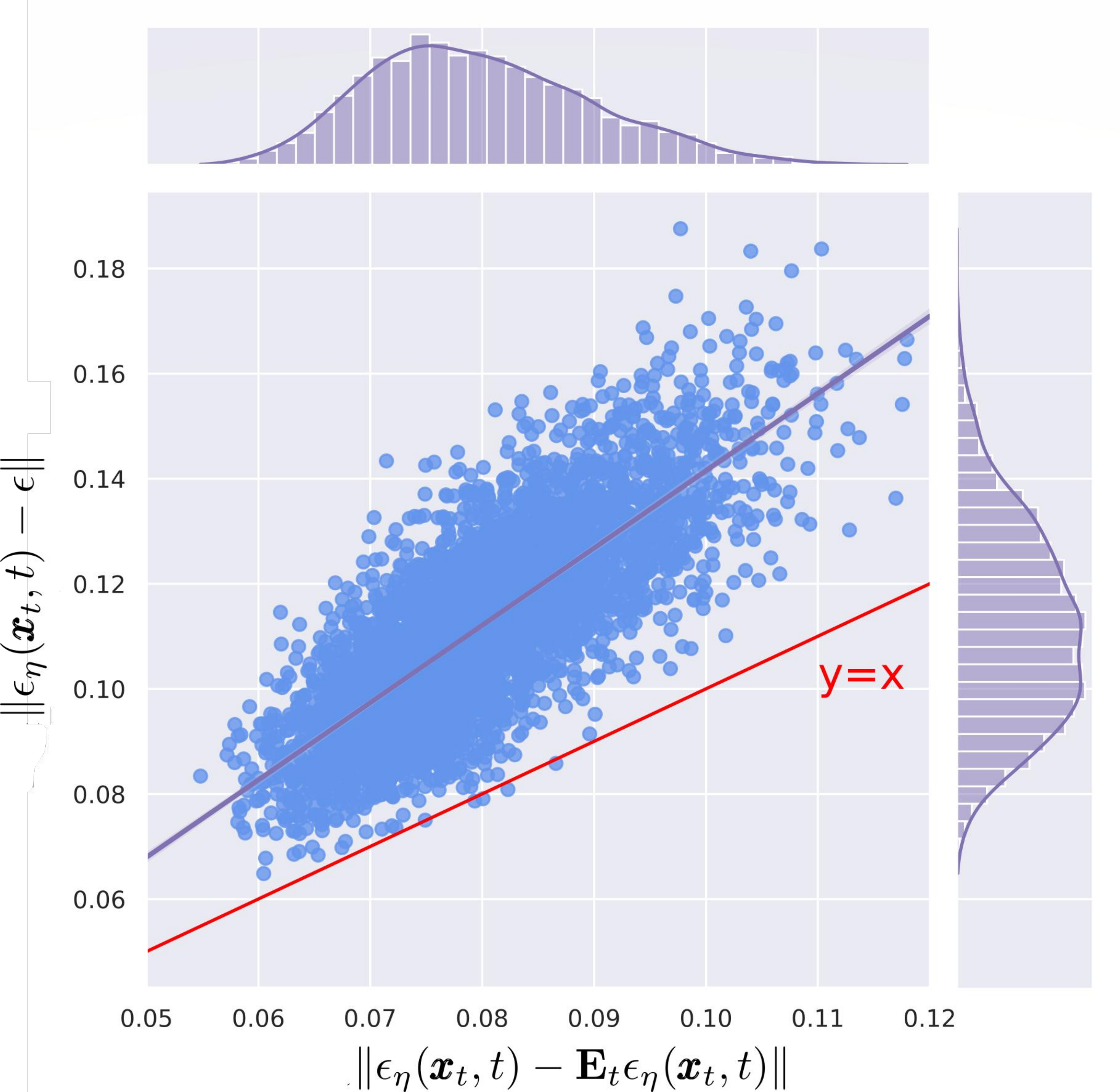}
        \caption{The correlation trend between the values of $\| \epsilon_\eta(\bm x_t, t) - \epsilon \|$ (fitting error) and $ \| \epsilon_\eta(\bm x_t, t)  - \mathbb{E}_t \epsilon_\eta(\bm x_t, t) \|$ (variance) at $t=399$ using $\epsilon$-prediction pre-trained diffusion model on ImageNet, given different $\bm x_0$'s and $\bm \epsilon$'s.}
        \label{fluctuation}
    \end{minipage}
        \hspace{.01in}
    \begin{minipage}[t]{0.32\linewidth}
        \centering
    \includegraphics[width=0.8\linewidth]{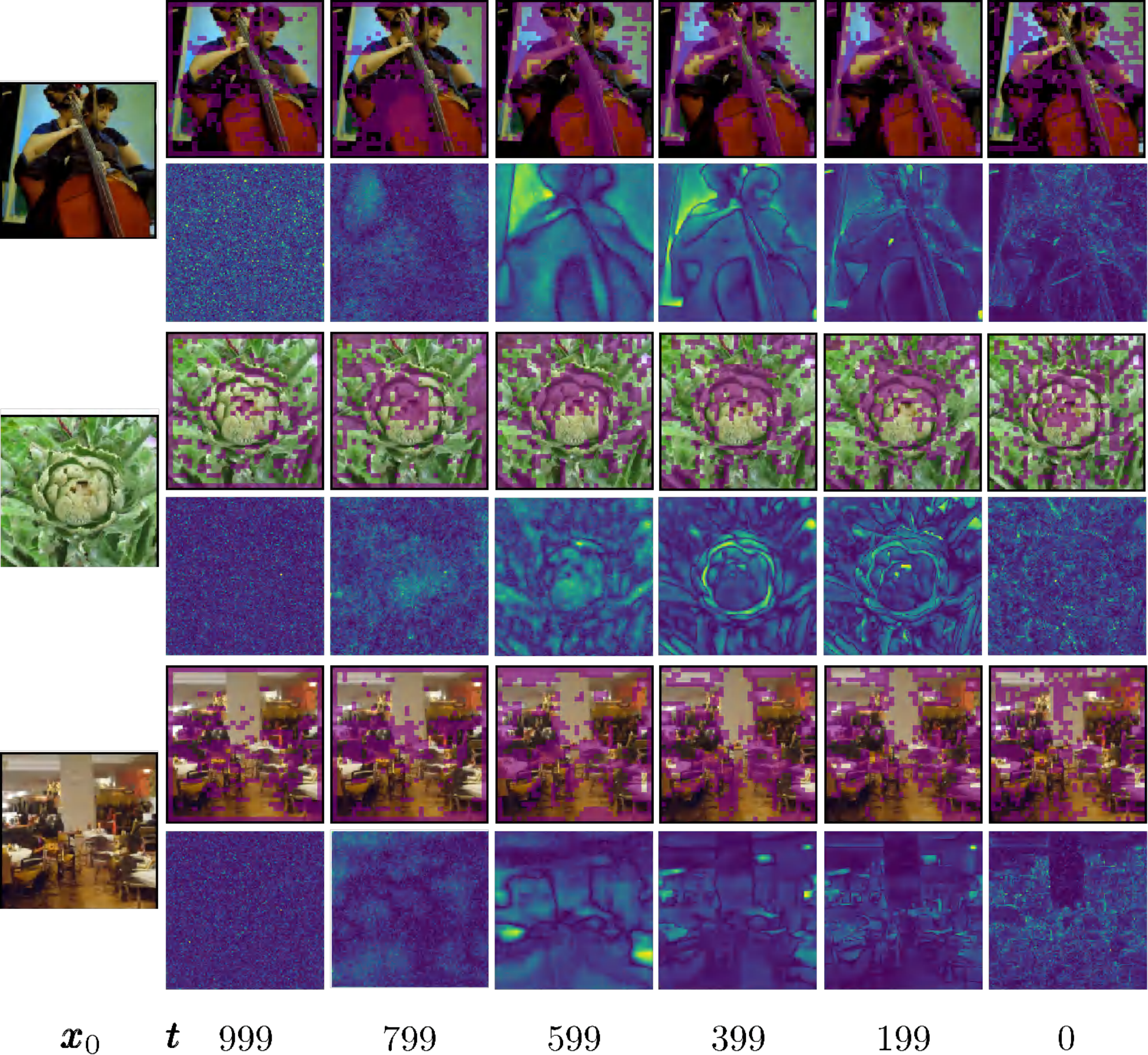}
        \caption{
        Visualization of attention maps and predicted noise variance on ImageNet diffusion. The left image in each group is the original image. The first row is the annotated attention maps while the second the noise variance at different $t$.
        }
        \label{attn_eps}
    \end{minipage}
    \hspace{.01in}
    \begin{minipage}[t]{0.32\linewidth}
        \centering
    \includegraphics[width=0.9\linewidth]{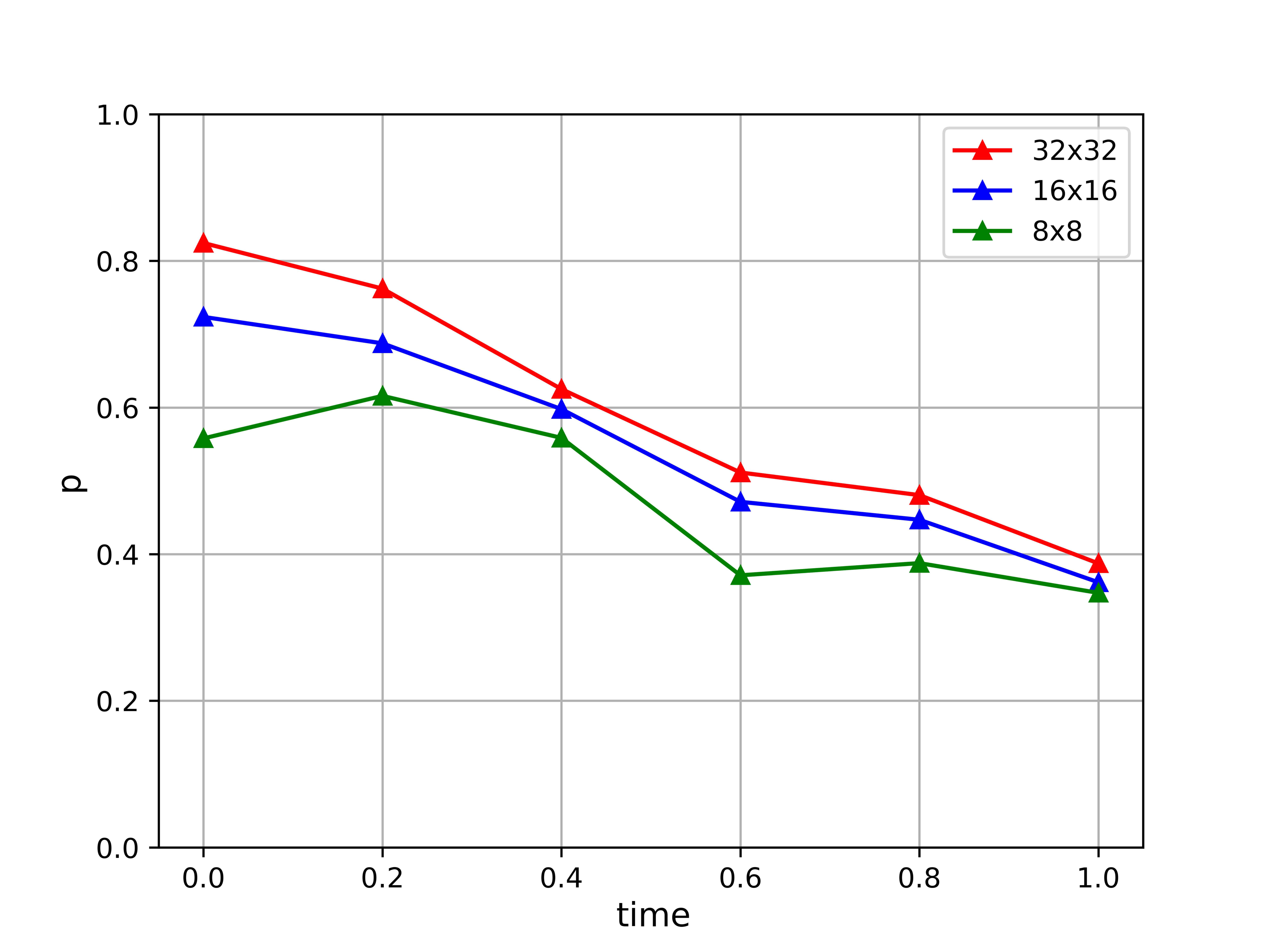}
        \caption{The Pearson correlation between the attention maps from different resolutions (8, 16, 32) and the predicted noise variance during sampling (1 $\rightarrow$ 0). Using $\epsilon$-prediction pre-trained diffusion model on ImageNet.}
        \label{p}
    \end{minipage}
\end{figure*}

\subsection{The Exploration of Reducing Fitting Error}
\lblsect{eps-and-attention}

\subsubsection{Reducing fitting error of the student} Figure \ref{cond_loss} shows the training loss of the conditional diffusion model is lower than that of the unconditional diffusion model. We also find that the conditional student model gives a denoised prediction $\bm{\hat x}_0$ closer to the teacher's prediction with better quality than the unconditional student during the sampling process (Figure \ref{cond_delta_x0}). These results suggest that semantic information (like labels) reduces the fitting error in the diffusion model. We will embed semantic information in a learned latent space for error reduction (\sect{DSE}).
 
\subsubsection{Reducing fitting error of the teacher} 
In this part, we present our findings that the fitting error is dominated by the prediction variation, and the variation is correlated to the self-attention map spatially.
Firstly, we perform a bias-variance decomposition of the $\epsilon$-prediction loss, which measures the fitting error (Eq.(\ref{eps_variance})). The first term (A) represents the prediction variance across the sampling procedure, while the second component (B) represents bias. 
With the training data, we estimate the fitting error and prediction variance values. As visualized in Figure \ref{fluctuation}, the fitting error is mainly determined by the variance spanning the entire global range, and the bias plays a minor role. 
Besides, the error is also positively correlated with the variance (see \app{analysis-eps} for more details).
Therefore, the fitting error can be largely reduced by restricting the prediction variance. Such insight conceptually coincides with the consistency assumption in Consistency Models~\cite{song2023consistency}.
\begin{equation}\label{eps_variance}
\begin{aligned}
\mathbb{E}_{\bm x_0,t,\epsilon}\| \epsilon_\eta(\bm x_t, t) - \epsilon \|
 =&  \underbrace{\mathbb{E}_{\bm x_0,t,\epsilon} \| \epsilon_\eta(\bm x_t, t)  - \mathbb{E}_t \epsilon_\eta(\bm x_t, t) \|}_{A} \\& +  \underbrace{\mathbb{E}_{\bm x_0,\epsilon} \|\mathbb{E}_t \epsilon_\eta(\bm x_t, t) - \epsilon \|}_{B}
\end{aligned}
\end{equation}


By analyzing the spatial configuration of the prediction variance, we find high variance tends to occur in regions with high self-attention scores. Inspired by~\cite{baranchuk2021label, tumanyan2022plug, kwon2022diffusion}, we opt to utilize attention modules from the decoder part of U-Net~\cite{ronneberger2015u} in the diffusion model to extract self-attention maps $\mathrm {A}_t^l$. which have been experimentally demonstrated to contain more representations. Specially, we perform global average pooling and upsampling on $\mathrm {A}_t^l$ to match the resolution of $\bm{\hat x}_t$.

As illustrated in Figure~\ref{attn_eps}, the predicted noise variance and the regions with high self-attention scores exhibit similar spatial distributions throughout the sampling process. 
By investigating the attention maps, we suggest that main subjects are mostly generated during the middle stages of sampling, while visual details are added during the final stages. 
This finding is supported by the experimental view of~\cite{baranchuk2021label, Wu2022UncoveringTD}. 
Consequently, a well-trained method should prioritize enhancing features in different granularity that require emphasis at various timesteps. 
We conduct Pearson correlation analysis between
the attention maps and the spatial prediction variance during the sampling process (Figure \ref{p}). 
As time $t$ goes from $1$ to $0$, the correlation between the variance and all attention maps at different resolutions gradually increases.
Notably, the attention score maps resolved at 32$\times$32 display the strongest positive correlation.
This finding implies the potential use of attention maps in error reduction.
\section{Method}
\lblsect{method}
Based on the observations in \sect{eps-and-attention}, we propose Spatial Fitting-Error Reduction Distillation  of denoising diffusion models (SFERD). SFERD uses attention guidance  and an extrinsic semantic gradient predictor to reduce the fitting error of the teacher and the student models, respectively. 
To better illustrate the improvements, we use DDPM process, DDIM sampler $g$ and $\bm \epsilon$-prediction pre-trained teacher model $\bm T_\eta$ by default in this section. 
Moreover, our methods can be extended to other samplers and predicted parameters including $v$-prediction or $x_0$-prediction~\cite{salimans2022progressive, ho2022imagen} easily.
\subsection{Teacher Model with Attention Guidance}
\lblsect{DAG}

\begin{figure}[ht]
    \centering
    \includegraphics[width = 1.01\linewidth]{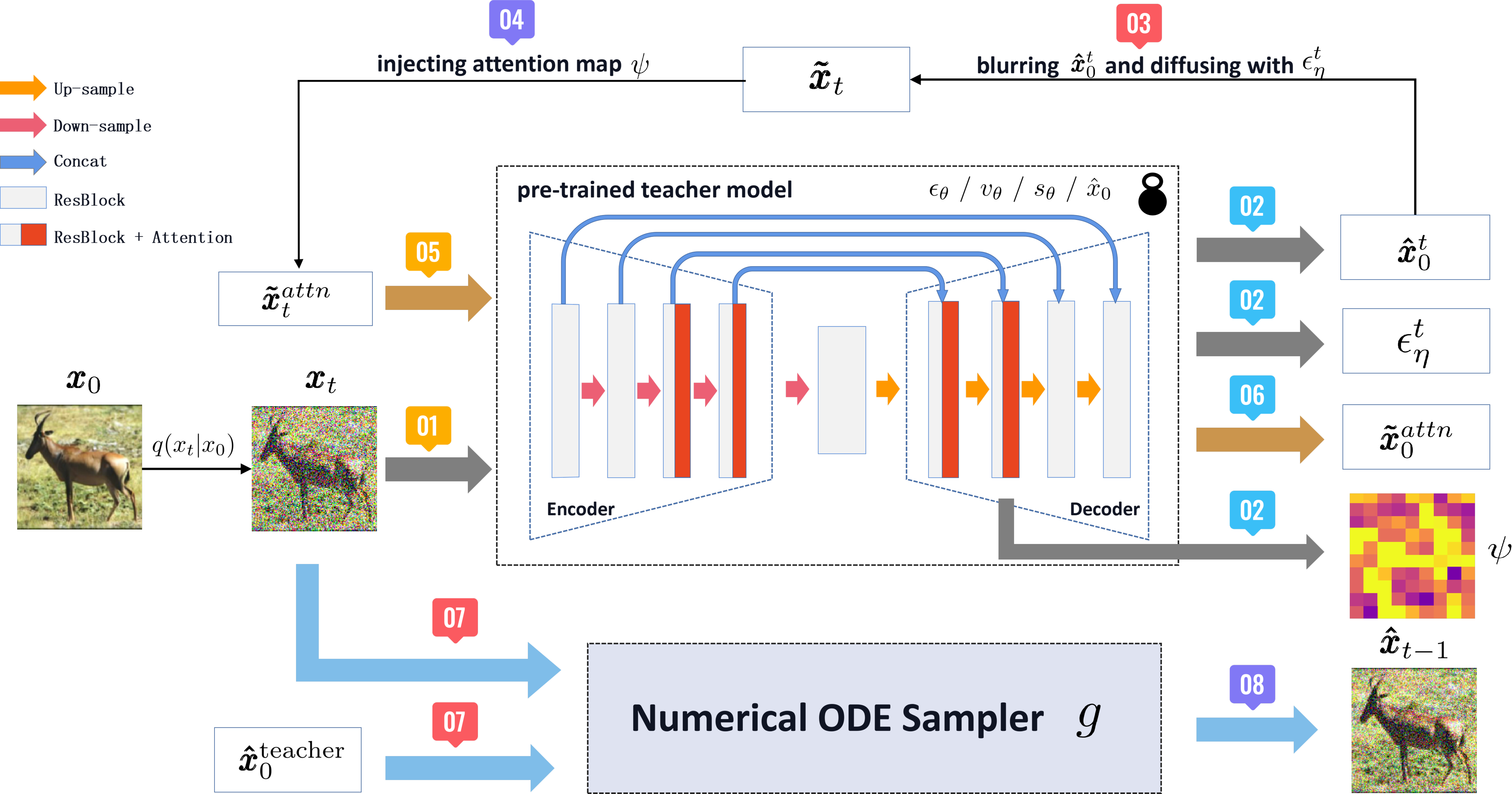}
    \caption{The illustration of attention guidance in the teacher model. Best viewed magnified on screen.}
    \label{DAG}
\end{figure}


Our approach focuses on identifying and optimizing high attention score regions (``Risky Regions''), which are strongly correlated with the teacher's fitting error in generated images. 
The process visualized in Figure~\ref{DAG} involves three key operations, including Gaussian blurring, attention injection and attention guidance sampling.

\subsubsection{Gaussian blurring} 
The real image $\bm x_0$ is first sampled by a forward process to obtain $\bm x_{t}$ (Step 1 in Figure \ref{DAG}). 
Then, given $\bm x_{t}$ and with reparameterization, the denoised prediction $\bm {\hat x_0^{t}}$ is predicted by the teacher model $\bm T_\eta$ at time $t$ (Step 2 in Figure \ref{DAG}). 
Next, we utilize Gaussian blur to introduce interference for the construction of unbalanced information. Specially, we deliberately destroy the Risky Regions from $\bm {\hat x_0^{t}}$ with Gaussian blur. This allows us to extract and optimize them later. 
Finally, we employ inverse DDIM to generate $\bm{\tilde{x}}_t = \sqrt{\bar{\alpha}_{t}}B(\bm{\hat{x}}_0^{t}) + \sqrt{1-\bar{\alpha}_{t}}\bm {\epsilon}_{\bm \eta}^{t}$ (Step 3 in Figure \ref{DAG}). $B(\cdot)$ denotes the Gaussian blur operation.

The main reason for using Gaussian blur is twofold. Conceptually, Gaussian blur can reduce trivial details in images while preserving the original manifold of the generated image.
Empirically, further experiments demonstrate in most time during sampling, the blurred denoised prediction is closer to the final generated sample than the original denoised prediction (\app{limitation}). 

\subsubsection{Attention injection} 
This operation is to further highlight the Risky Regions since $\bm {\tilde x}_{t}$ after Gaussian blurring is globally blurred instead of being applied to the region only.
Let $f_{t}^{l}$ denote the hidden feature fed to the attention blocks in layer $l$ at time $t$.
After $N$ heads of self-attention blocks, the output self-attention map $A_{t}^l$ can be expressed as:
\begin{equation}\label{attn_map}
\begin{aligned}
     A_{t}^l = \text {softmax}(Q^{l,a}_{t}(K^{l,a}_{t})^T/\sqrt{d}) \\ \mathrm{where} \quad
     Q^{l,a}_{t}  = f_{t}^l W_Q^{l,a} \,,  K^{l,a}_{t}  = f_{t}^l W_K^{l,a}
\end{aligned}
\end{equation}
Here $a \in [0, N-1]$ denotes an attention head, and $d$ is the output dimension of queries $Q^{l,a}_{t}$ and keys $K^{l,a}_{t}$. After that, we upsample $A_{t}^l$ to the image size and extract the regions with high attention scores with Eq.(\ref{attn_injection}).
\begin{equation}\label{attn_injection}
\begin{aligned}
     \psi = \mathbb{I} (A_{t}^l>k) \,\,, \quad 
     \bm{\tilde x}_{t}^{attn} = (1-\psi)\odot \bm x_{t} + \psi \odot \bm{\tilde x}_{t}
\end{aligned}
\end{equation}
Here $\psi$ denotes a Boolean matrix where a pixel is $1$ if its attention value is over a given threshold $k$ and $0$ otherwise. 
$\odot$ denotes the Hadamard product. 
$\bm{\tilde x}_t^{attn}$ is identical to $\bm x_t$ in regions with low self-attention scores but becomes blurred in regions with high scores (Step 4 in Figure \ref{DAG}).

\subsubsection{Attention guidance sampling}
In this operation, the denoised prediction $\bm {\tilde x}_{0}^{attn}$ is calculated from the teacher model with $\bm {\tilde x}_t^{attn}$
(Steps 5, 6 in Figure \ref{DAG}).
Together with $\bm {\hat x}_0^{t}$, the improved denoised prediction $\bm {\tilde x}_0^{\mathrm {teacher}}$ can be obtained. Both calculations are in Eq.(\ref{x_0_teacher}). Subsequently, the DDIM sampler conditioned on $\bm x_t$ and $\bm {\tilde x}_0^{\mathrm {teacher}}$ is applied to get $\bm {\hat x}_{t-1}$ (Step 7, 8 in Figure \ref{DAG}). Repeating the above three operations until $\bm{\hat x}_{t_i}$ is obtained. Finally, we get $\tilde{\bm x}_0^{\mathrm{target}}$ by Eq.(\ref{target-function}).
\begin{equation}\label{x_0_teacher}
    \begin{aligned}
    \bm {\tilde x}_0^{attn} &=\frac{\bm {\tilde x}_t^{attn} - \sqrt{1-\bar{\alpha}_{t}}\bm {\epsilon}_{\bm \eta}(\bm {\tilde x}_t^{attn}, t)}{\sqrt{\bar{\alpha}_{t}}} \\
        \bm {\tilde x}_0^{\mathrm {teacher}} &=
        \bm {\hat x}_0^{t} + w \times(\bm{\hat x}_0^t - \bm {\hat x}_0^{attn})
    \end{aligned}
\end{equation}
Here $w$ denotes the attention guidance strength.
$(\bm{\hat x}_0^t - \bm {\hat x}_0^{attn})$ contains the semantic information differences for guidance highlighted by the high attention scores, helping the teacher to improve the quality of denoised prediction. 
The approach relies entirely on the diffusion model's intrinsic representations and provides a new perspective of guidance. 
It is unsupervised and still supports the extension of incorporating external conditions for better performance. 

 \begin{table*}[ht]
\centering
\begin{tabular}{lrrrrr}
\noalign{\smallskip}\noalign{\smallskip}\toprule
\multirow{2}{*}{Model} & \multirow{2}{*}{NFE} &\multicolumn{2}{c}{CIFAR-10 32$\times$32} &\multicolumn{2}{c}{ImageNet 64$\times$64}\\
 & & FID ($\downarrow$) & IS ($\uparrow$) &{FID ($\downarrow$)} & IS ($\uparrow$)\\
  \midrule
 SFERD-PD (ours) & 1 &7.54 &8.61 &14.85 & 36.55  \\ 
 PD~\cite{salimans2022progressive} & 1 & 14.85 & 7.96 & 19.23 & 22.73   \\
   SFERD-CD (ours)& 1 &\textbf{5.31} &\textbf{9.24} & \textbf{9.39}& \textbf{47.19} \\ 
 CD~\cite{song2023consistency} & 1 & 8.21 & 8.41 & 13.87 & 29.98   \\
 $\mathrm{DFNO}^{*}$~\cite{zheng2023fast} & 1 & 4.12 & \multicolumn{1}{c}{/} & 8.35 & \multicolumn{1}{c}{/}   \\
  ViTGAN~\cite{leevitgan} & 1 & 6.66 & 9.30 & \multicolumn{1}{c}{/} & \multicolumn{1}{c}{/}   \\
DiffAugment-BigGAN~\cite{zhao2020differentiable} & 1 & 5.61 & 9.16 &  \multicolumn{1}{c}{/} & \multicolumn{1}{c}{/}  \\
    TransGAN~\cite{jiang2021transgan} & 1 & 9.26 & 9.02 & \multicolumn{1}{c}{/} & \multicolumn{1}{c}{/}   \\
 \midrule
 SFERD-PD (ours) & 2 & 6.37& 8.92&7.53 &45.72   \\
 PD~\cite{salimans2022progressive} & 2 & 7.64 & 8.85 & 9.71 & 25.37  \\
 SFERD-CD (ours) & 2 &\textbf{4.19} &\textbf{9.45} &\textbf{6.08} &\textbf{54.05}  \\
 CD~\cite{song2023consistency} & 2 & 6.26 & 9.17 & 8.24 & 32.60  \\
  \midrule
 SFERD-PD (ours) & 4 & 3.44 & 9.32 &5.93 &55.19 \\
 PD~\cite{salimans2022progressive} & 4 & 4.28 & 9.25 & 7.22 & 30.72  \\
  SFERD-CD (ours) & 4 & \textbf{2.68} & \textbf{9.79} &\textbf{4.41} &\textbf{57.98}\\
 CD~\cite{song2023consistency} & 4 & 3.39 & 9.71 & 5.81 & 35.41  \\
  \midrule
   SFERD-EDM (ours)& 35 & \textbf{2.12} & \textbf{9.88} & \multicolumn{1}{c}{/} & \multicolumn{1}{c}{/}\\
   EDM~\cite{karras2022elucidating} & 35 & 2.41 & 9.83 & \multicolumn{1}{c}{/} & \multicolumn{1}{c}{/}\\
     \midrule
 SFERD-EDM (ours)& 79 & \multicolumn{1}{c}{/}& \multicolumn{1}{c}{/} & \textbf{2.43} & \textbf{74.73}\\
  EDM~\cite{karras2022elucidating} & 79 & \multicolumn{1}{c}{/} & \multicolumn{1}{c}{/} & 2.99 & 39.09\\
   \midrule
 SFERD-DDIM (ours) & 1024 & 2.27 & 9.80 &2.82 & 69.17 \\
 DDIM~\cite{song2020denoising} & 1024 & 2.58 & 9.76 &3.34 & 37.55 \\

\bottomrule
\end{tabular}
\vspace{3pt}
\caption{Sample quality on CIFAR-10 and ImageNet 64$\times$64. SFERD-* represents the implementation of the corresponding model within  SFERD framework. For example, SFERD-PD and SFERD-RD represent the models that refer to ideas of Progressive Distillation (PD) and Consistency Distillation (CD) and imply within SFERD, respectively. Both attention guidance method and semantic encoding-based gradient predictor are introduced.}
\label{tab:i64_cifar10}
\end{table*}
\subsection{Student Model with Semantic Gradient Predictor}
\lblsect{DSE}
\begin{figure}[ht]
    \centering
\includegraphics[width=0.92\linewidth]{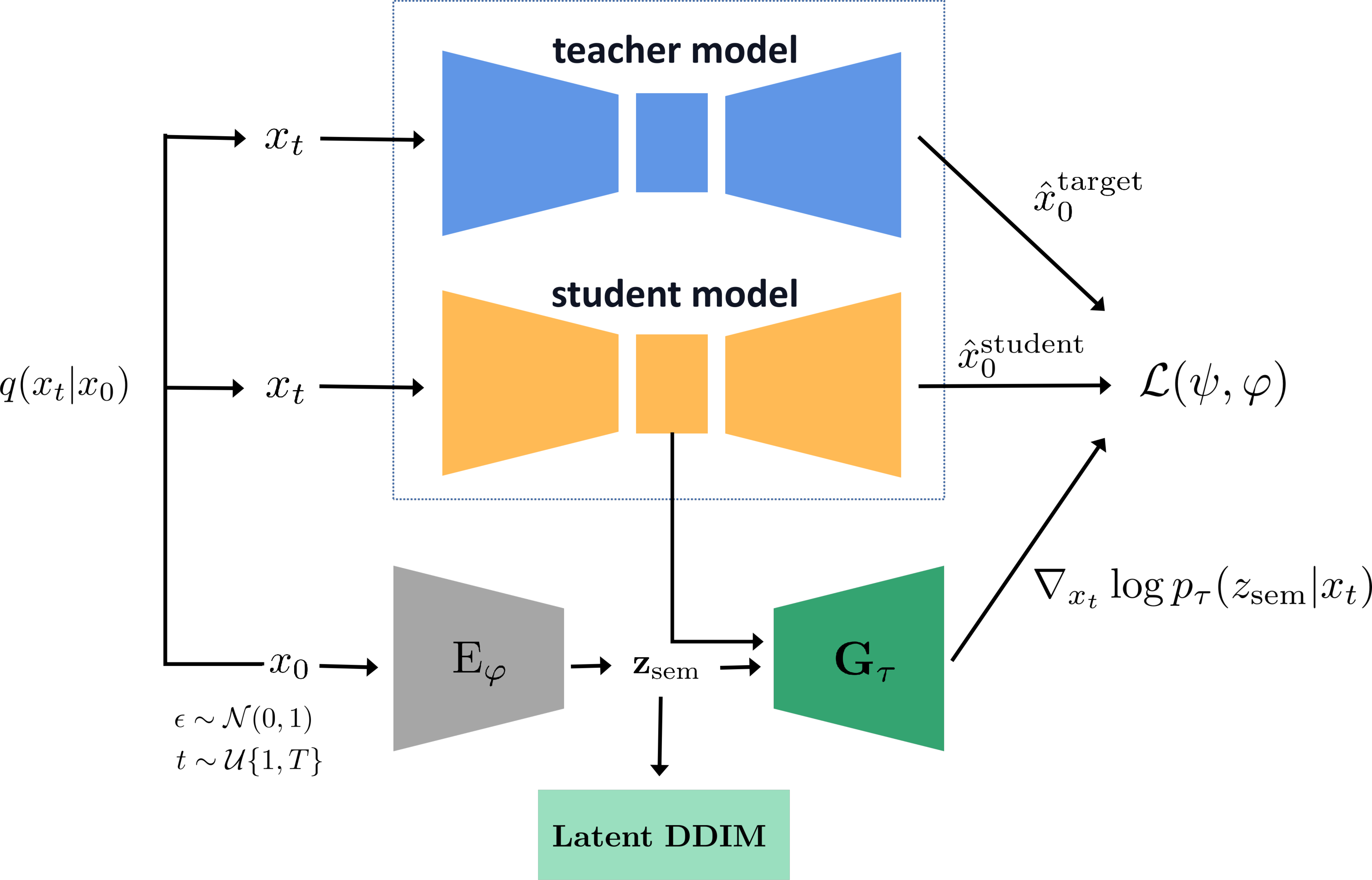}
    \caption{The training process of distillation with semantic gradient predictor. Note that the original student model is trained.}
    \label{DSE}
\end{figure}

The gradients with semantic information calculated by the classifier have been validated for their ability to compensate for information bias and loss that arise during sampling~\cite{dhariwal2021diffusion}. 
To reduce the fitting error of the trained distillation student model, we introduce a learned semantic encoder in the student model, which provides a latent vector containing more intact reconstruction information.
In detail, we integrate a semantic encoder $\bm z_{\mathrm {sem}} = \mathrm{E_\varphi}(\bm x_0)$ and a predictor $G_\tau(\bm x_t,\bm z_{\mathrm {sem}},t)$ for the student to learn representations from the real image $x_0$. 
This would help in fitting to $\bm{\tilde x}_0^{\mathrm {target}}$. 
Previous work~\cite{dhariwal2021diffusion} introduce an extra label $y$ in the conditional diffusion model.
When we replace class label conditions $\bm y$ with the learned latent equation $\bm z_{\mathrm {sem}}$:
\begin{equation}
\label{cond_p}
\begin{aligned}
        p_{\theta,\varphi}(\bm{x}_{t-1} | \bm{x}_{t}, \bm{z}_{\mathrm {sem}}) &\approx \mathcal{N}(\bm{x}_{t-1}; \bm{\mu}_{\theta}(\bm{x}_{t},t) + \bm{\Sigma}_{\theta}(\bm{x}_{t}, t) \cdot \\& \qquad \nabla_{\bm{x}_{t}} \log p_{\tau}(\bm{z}_{\mathrm {sem}} | \bm{x}_{t}) \,,\, \bm{\Sigma}_{\theta}(\bm{x}_{t}, t)) \, 
\end{aligned}
\end{equation}
Based on Eq.(\ref{cond_p}), the training objective of the distillation model can be reformulated as Eq.(\ref{gradient_predictor}). 
\begin{align}
    & \mathcal{L}(\theta, \varphi, \tau) 
    \notag 
     \\ =&
    \mathbb{E}_{\bm{x}_{0},t,\epsilon}\bigg[\omega(\lambda_t)\big\| \bm{\Sigma}_{\theta}(\bm{x}_{t}, t) \cdot \nabla_{\bm{x}_{t}} \log p_{\tau}(\bm{z}_{\mathrm {sem}}|\bm{x}_t) \notag \\ & - \big(\widetilde{\bm{\mu}}_{t}(\bm{x}_{t},\bm{\tilde x}_{0}^{\mathrm{target}}) - \bm{\mu}_{\theta}(\bm{x}_{t},t)\big)  \big\|^{2}\bigg] 
      \\ \label{gradient_predictor} =&
     \mathbb{E}_{\bm{x}_{0},t,\epsilon}\bigg[\omega(\lambda_t)\big\| 
     \frac{(1-\bar \alpha_t)\bm{\Sigma}_{\theta}(\bm{x}_{t}, t)}{\sqrt{\bar \alpha_{t-1}}\beta_t}  \cdot \nabla_{\bm{x}_{t}} \log p_{\tau} (\bm{z}_{\mathrm {sem}}|\bm{x}_t) \notag \\ &  -
    \big(\bm{\tilde x}_0^{\mathrm {target}} - \bm{\hat x}_0^{\mathrm {student}}(\bm{x}_{t},t)\big) \big\|^{2}\bigg]
\end{align}
where $\bm{\Sigma}_{\theta}=\sigma_t^2\bm{\mathrm{I}}=\frac{1-\bar{\alpha}_{t-1}}{1-\bar{\alpha}_{t}}\beta_{t}\bm{\mathrm{I}}$. Eq.(\ref{gradient_predictor}) uses the gradient $\nabla_{\bm{x}_{t}} \log p_{\tau}(\bm{z}_{\mathrm {sem}}|\bm{x}_t)$ with semantic information to compensate for the fitting error with $\bm{\tilde x}_0^{\mathrm {target}}$. 
We design a predictor $G_\tau(\bm x_t,\bm z_{\mathrm {sem}},t)$ to approximate $\nabla_{\bm{x}_{t}} \log p_{\tau}(\bm{z}_{\mathrm {sem}}|\bm{x}_t)$. 
Figure \ref{DSE} shows the training network and data flow. 
By default, $G_\tau$ is not trained directly to fit $\nabla_{\bm{x}_{t}} \log p_{\tau}(\bm{z}_{\mathrm {sem}}|\bm{x}_t)$ but combined with the distillation model.
The trained student model is frozen during the first half of the training epochs and jointly optimized with $G_\tau$ and $E_\varphi$ using a low learning rate in the latter half.
We find in this case the optimized $G_\tau$ produces gradients close to $\nabla_{\bm{x}_{t}} \log p_{\tau}(\bm{z}_{\mathrm {sem}}|\bm{x}_t)$. 
We validate the crucial role of incorporating $\bm{z}_{\mathrm {sem}}$ in ensuring the effectiveness of the improved method through ablation experiments. 
The student model can be combined with  $G_\tau$ to perform a few-step sampling using the formula similar to the classifier guidance.
In addition, we also train another DDIM sampler $p_\omega(\bm z_{\mathrm {sem}}^{t-1}|\bm z_{\mathrm {sem}}^t)$ to sample $\bm z_{\mathrm {sem}}$ in the latent space, 
following the approach as \cite{preechakul2022diffusion}. 
Both $\mathrm{E_\varphi}$ and $G_\tau$ are independent of the original distillation model and can be easily integrated into the trained distillation process, achieving less training time. 
The implementation details and pseudo code in this part are provided in \app{algorithm}, B.

\section{Related Work}

The goal of our methods is to accelerate sampling while maintaining high image quality, in line with many existing works. For instance, Watson et.al (2021, 2022) exploit using traditional dynamic programming methods to accelerate sampling, Zhang et al (2022) proposes the use of an exponential integrator (DEIS) to speed up sampling. The use of knowledge distillation in diffusion models can be traced back as far as the DDIM-based one-step denoising model (DS) implemented in~\cite{luhman2021knowledge}, whose main drawback is that the full steps of the original teacher model have to be fully applied for each training. The Classifier-Guided Distillation (CFD) proposed in~\cite{sun2022accelerating} uses a classifier to extract the sharpened feature distribution of the teacher into the students and uses the KL divergence as training loss, allowing it to focus on the focal image features. Higher-Order Denoising Diffusion Solvers (GENIE)~\cite{dockhorn2022genie}, on the other hand, achieve accelerated sampling by adding a prediction module capable of receiving information from distillation Higher Order Solvers to the network backbone. Progressive distillation (PD)~\cite{salimans2022progressive} and guided distillation (GD)~\cite{meng2022distillation}  implement unconditional and conditional distillation diffusion models, respectively, at the cost of halving the number of sampling steps per iteration, in addition to GD which extends the training to the latent space and implements random sampling. Consistency Distillation (CD)~\cite{song2023consistency} exploits the self-consistency of the ODE generation process by minimizing the difference between two noisy data points on the same ODE path to achieve few-step distillation.
Our two improvement methods can be applied to the majority of diffusion distillation models above. Moreover, to the best of our knowledge, our study is the first to introduce the semantic encoder into the distillation process of the diffusion model.

\section{Experiments}
\lblsect{experiments}
\subsection{Experimental Setting}
We mainly examine SFERD on two standard image generation benchmarks, including CIFAR-10 and class-conditional ImageNet 64$\times$64. We measure the performance using Fr\'echet Inception Distance (FID)~\cite{heusel2017gans} and Inception Score (IS)~\cite{salimans2016improved}. 
All results are computed from 50,000 sampled generated images and averaged over 3 random seeds. All students are uniformly initialized by the corresponding teachers.  Additional experimental details are provided in \app{implementation-details}.
\subsection{Few-steps Image Generation}
Our distillation framework allows different configurations of diffusion process (like Variance Preserving or Exploding~\cite{song2020score}), numerical solver, total diffusion timesteps and so on.
For fair 
comparisons, we choose to implement SFERD by referring to the ideas of Progressive Distillation models (PD)~\cite{salimans2022progressive} and Consistency Distillation models (CD)~\cite{song2023consistency}. Specifically, PD can be interpreted as setting the target function $\bm {R}$ in SFERD to $\bm T_\eta$,
while $\bm {R}$ is set to the student model $\bm S_{\theta^-}$ updated by the exponential moving average (EMA) in CD. 
In our experiments, PD is applied to the teacher network from unconditional ADM~\cite{dhariwal2021diffusion} using DDPM noise schedule and DDIM (Euler) sampler, and CD is applied to the teacher from EDM~\cite{karras2022elucidating} and $2^{nd}$ Heun sampler. We unify the metric function to $\ell_2$ distance. 
We pretrain all teachers using the configurations specified in the original papers. The initial teacher model of PD is set to 1024 sampling steps, whereas for CD, it is set to 18 or 40 steps. By default, 
the students of the previous distillation stage serve as the teachers for the next stage in SFERD-PDs.

We compare our SFERD with PD and CD on the CIFAR-10 and ImageNet 64$\times$64. Results in Table~\ref{tab:i64_cifar10} demonstrate that the performance of all models improved as the sampling steps increased. Notably, SFERD-PD and SFERD-CD achieved better performance than their baseline models, and SFERD-CD displays superior performance across all datasets and sampling steps. 
Our improved methods can be applied not only to mainstream diffusion distillation models but also to enhance the performance of pre-trained models directly through fine-tuning. Specifically, we apply our methods on pre-trained unconditional ADM and EDM directly, aligning the distillation steps of the student with that of the teacher, which can be easily achieved by setting $s=t-1$. The results shown in Table~\ref{tab:i64_cifar10} indicate superior performances of ADM and EDM, which are improved through the integration of attention guidance and semantic gradient predictor.
Additional sample examples are available in \app{additional-samples}.

\subsection{Ablation Studies}
\lblsect{ablation-studies}
We conduct ablation experiments on the design of critical hyperparameters in the training of SFERD. All ablations are performed on the conditional ImageNet 64×64 using SFERD-PD (no semantic gradient predictor in the student) with 4 sampling steps unless otherwise stated. 

\begin{figure}[t]
    \centering
    \begin{subfigure}[b]{0.45\linewidth}
        \includegraphics[width = \linewidth]{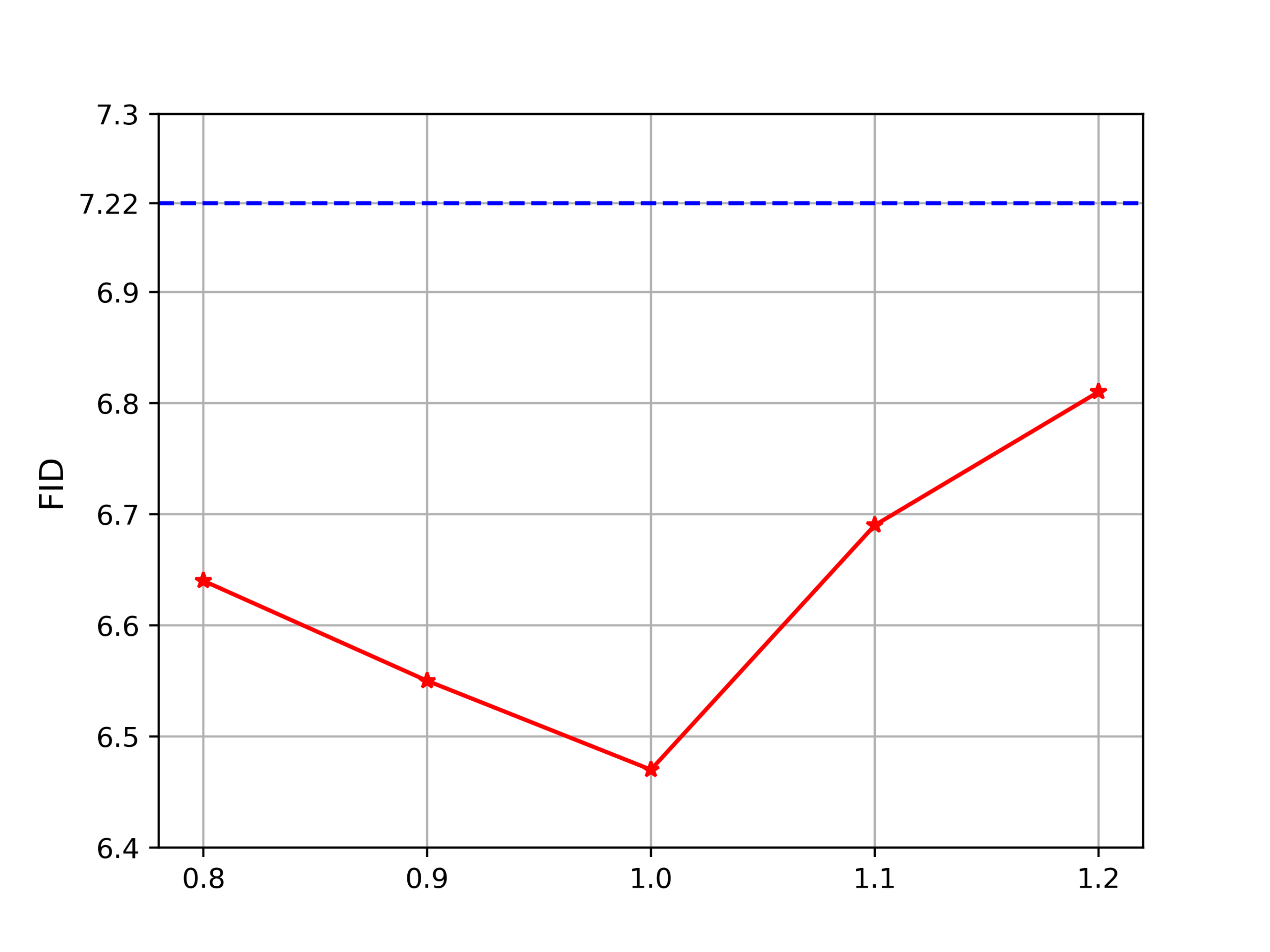}
    \caption{\textbf{Attention threshold ($\psi$).}}
    \label{attn_threshold}
    \end{subfigure}
    \hspace{.01in}
    \begin{subfigure}[b]{0.45\linewidth}
    \includegraphics[width=\textwidth]{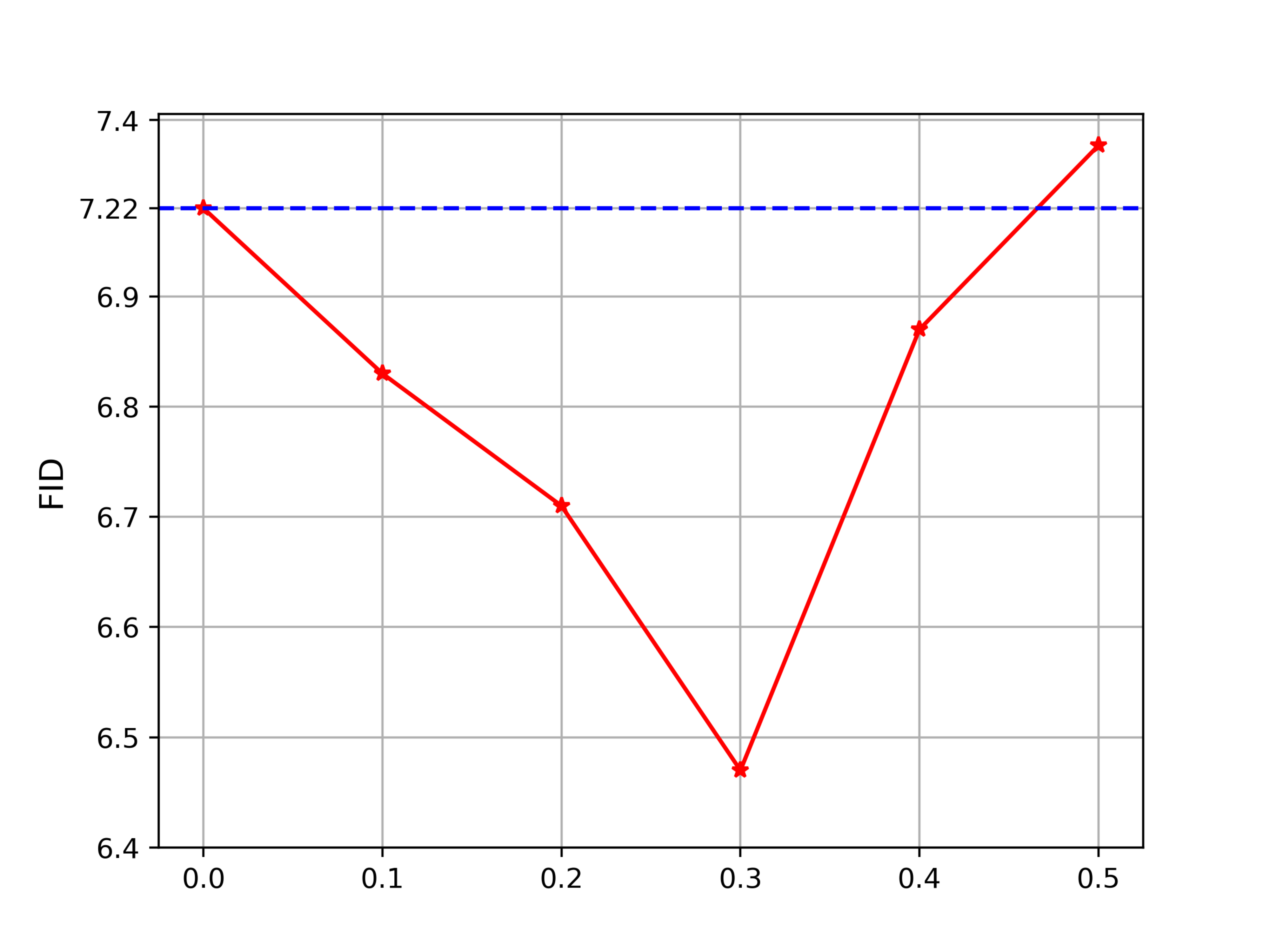}
        \caption{\textbf{Attention guidance strength ($\omega$).}}
        \label{attn_guidance}
    \end{subfigure}%
    \caption{The ablation studies of various factors which affect SFERD on ImageNet 64x64. The studies are done on SFERD-PD. To verify the effectiveness of attention guidance, SFERD-PD used in ablations does not apply semantic encoding-base gradient predictor to the student when training. The blue line represents the baseline, which is FID of the original PD with 4 sampling steps. The best configuration for SFERD-PD is $\sigma =3 \,\, ,\psi=1.0$, and $w=0.3$. The rest visualizations are presented in SM.}
\end{figure}

\begin{figure}[t]
    \centering
\includegraphics[width=1.01\linewidth]{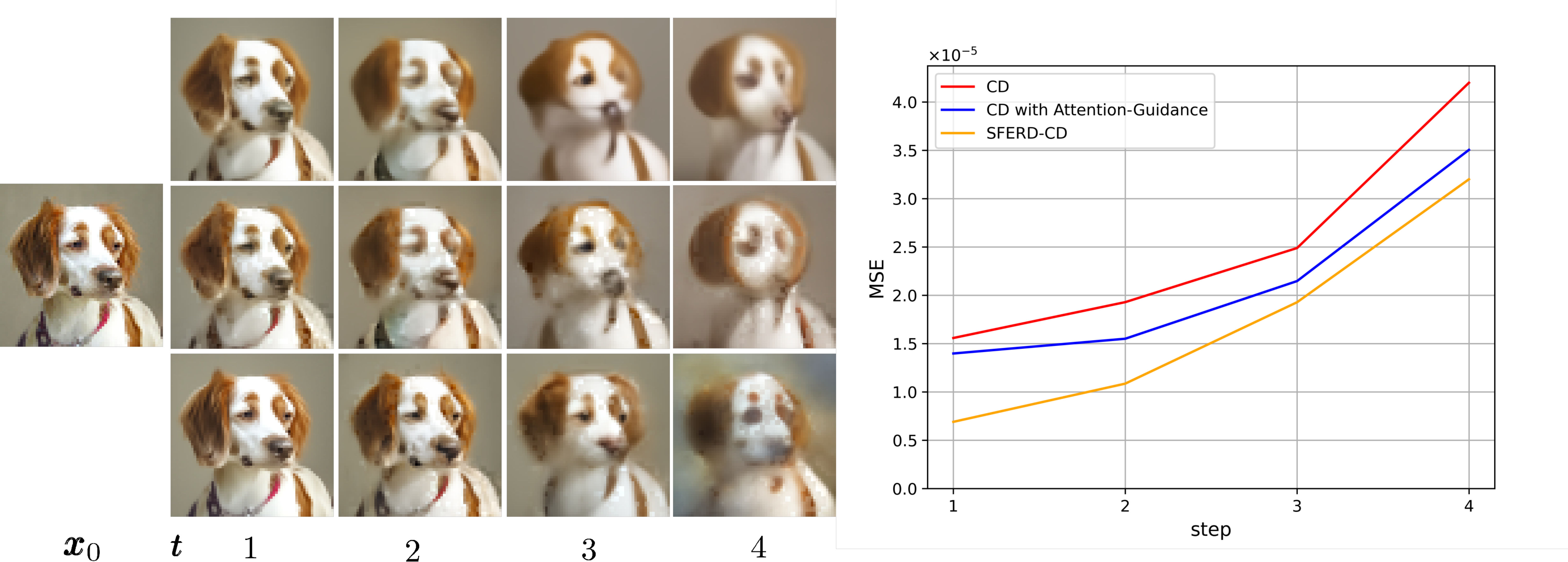}
    \caption{Left: The images of denoised prediction $\bm \hat{x}_0^t$ by denoising $\bm x_t$ at time $t$. The first-row use CD, the second-row use CD with attention guidance method and the third-row use SFERD-CD. All models are sampled with 4 steps. Right: Average $\ell_2$ for all timesteps.}
    \label{validation}
\end{figure}


\subsubsection{Attention threshold.} In order to determine the attention threshold, we compute the scales of 0.8, 0.9, 1.0, 1.1 and 1.2. As shown in Figure \ref{attn_threshold}, the best metrics are obtained when the $\psi$ is 1.0. A threshold that is too high or too low can deteriorate training performance.
\subsubsection{Attention guidance strength.} We evaluate the effect of attention guidance strength on performance (Figure \ref{attn_guidance}), calculating the scales from $0$ to $0.5$. The best FID was achieved when $w=0.3$.

\subsubsection{Gaussian blur strength.} We evaluate the effect of Gaussian blur strength $\sigma$ on performance. We test the strength values of 1, 3, 5, and obtain the best FID at $\sigma$ = 3.
\subsubsection{Denoising ability.} Specially, we randomly select 500 real images $\bm x_0$ from ImageNet 64$\times$64 and perform forward diffusion on them to obtain $\bm x_t$. We then use pre-trained 4-sampling-step CD, CD with attention guidance and SFERD-CD to predict the denoised image $\bm {\hat x}_0^t$ for a single step at $t$. Figure \ref{validation} (left) shows $\bm {\hat x}_0^t$ generated by either CD with attention guidance alone or SFERD are more precise and realistic than those generated by CD at the corresponding timesteps. Furthermore, we compute the 
$\ell_2$ distances between $\bm x_0$ and $\bm {\hat x}_0^t$ generated by three models, averaging over 500 random images. Figure \ref{validation} (right) shows the results, indicating that both enhancements of SFERD obviously reduce the $\ell_2$ distances of the denoised prediction at each timestep. Moreover, the FID of CD, CD with attention guidance, and SFERD-CD are 
presented in order: 5.81, 5.27, and 4.41, respectively.
\section{Conclusion}

In conclusion, we propose the Spatial Fitting-Error Reduction Distillation model (SFERD) for Denoising Diffusion Models. SFERD effectively enhances performance using intrinsic and extrinsic representations, generating high-quality samples in only few steps. The core idea behind SFERD is to reduce the fitting error of the student and the teacher in distillation. It is achieved through the dependent use of attention guidance we proposed for the teacher and an external semantic gradient predictor in the student model. 
\section{Acknowledgements}
This paper is funded by National Key R$\&$D Program of China (2022YFB3303301), National Natural Science Foundation of China (Grant No. 62006208), and Youth Program of Humanities and Social Sciences of the Ministry of Education (No.23YJCZH338). This paper is also supported by Alibaba-Zhejiang University Joint Research Institute of Frontier Technologies. We are grateful to Jionghao Bai and Jingran Luo for supplementing the appendix. We also thank to Chenye Meng and Haoran Xu for their helpful comments and discussion, and to Jiahui Zhang, Qi Liu, Ying Zhang, and Yibo Zhao for proofreading the draft. 

{
\small
\bibliography{aaai24}
}

\newpage
\clearpage
\appendix


\section{Algorithm}
\lblapp{algorithm}
In this section, we provide the algorithms for the training and sampling of our SFERD model. 
We define the distribution of the forward diffusion process conditioned on $\bm x_0$ as $q(\bm x_t|\bm x_0) = \mathcal{N}(\bm x_t|\mu_t \bm x_0, \sigma_t \mathbf{I})$, 
where $\mu_t, \sigma_t \in \mathbb{R}^{+}$ are differentiable functions of t with bounded derivatives.

\renewcommand{\algorithmiccomment}[1]{\bgroup\quad\quad//~#1\egroup}
\begin{algorithm}[h]
\centering
\caption{SFERD Training}\label{alg:msrd_training}
\begin{algorithmic}[1]
    \REQUIRE Data set $\mathcal{D}$
    \REQUIRE Trained teacher model $\bm T_\eta$\,\,, Student model $\bm S_\theta$\,, Target function $\bm R$
    \REQUIRE Loss weight function $\omega(\cdot)$
    \,, Masking threshold  $ k $
    , Distillation speed  $ n $
    \REQUIRE Semantic encoder $\mathrm{\bm E_\varphi}$\,\,, Gradient predictor $\bm G_{\tau}$
    \WHILE{not converged}
        \STATE $\bm x_0 \sim \mathcal{D}$ \COMMENT{Sample data}
        \STATE $\bm \epsilon \sim \mathcal{N}(0, \mathbf{I})$ \COMMENT{Sample noise}
        \STATE $s \sim \{ 0, n, 2n,...,T-2n,T-n \}$
        \STATE $i \sim \{1,...,n\}$
        \STATE $t = s+i$ \COMMENT{Sample time}
        \STATE $\bm x_{t} = \mu_t \bm x_0 + \sigma_t \bm \epsilon$ \COMMENT{Add noise to data}
        \STATE $\bm \epsilon_{\bm \eta}^{t}, \bm A_{t}^l = \bm x_{\bm \eta}^\mathrm {teacher}(\bm{x}_{t}, t)$ \COMMENT{Get predicted outputs}
        \STATE $\bm {\hat x}_0^{t} = (\bm{x}_{t}-\sigma_t\bm \epsilon_{\bm \eta}^{t})/\mu_t$\COMMENT{Re-parameterisation}
        \STATE $\psi = \mathbb{I} (A_{t}^l>k)$ \COMMENT{generate masked attention map}
        \STATE $\bm{\tilde{x}}_t = \mu_t\textrm{Gaussian\text-Blur}(\bm{\hat{x}}_0^{t}) + \sigma_t\bm {\epsilon}_{\bm \eta}^{t} $\COMMENT{Gaussian blurring and DDIM-inverse}
        \STATE $\bm{{\tilde{x}}}_{t}^{attn} = (1-\psi) \odot {\bm{x}}_{t}+\psi \odot \bm{\tilde{x}}_{t}$\COMMENT{Attention injection}
        \STATE $\bm {\tilde x}_0^{\mathrm  teacher}=(1+w)\bm{\hat x}_0^{t}-w \bm {\hat x}_0^{attn}$ \COMMENT{Compute teacher predicted images}
        \STATE $\lambda_t = \log(\mu_{t}^2 /\sigma_{t}^2)$ \COMMENT{log-SNR}
        \STATE $\bm {\hat x}_{t-1}=g(\bm x_t, \bm {\tilde x}_0^{\mathrm teacher})$  \COMMENT{Teacher Sampling}
        \STATE $\bm{\tilde x}_0^{\mathrm target}=\bm R(\hat {\bm x}_{t-1}, s)$ \COMMENT{Compute target value}
        \STATE $z_\mathrm{sem} = E_\varphi(\bm x_0)$ \COMMENT{Obtain latent vector}
        \STATE $L_{{\bm\theta}} = \omega(\lambda_t) \lVert \frac{\sigma_{t-1}^2}{\mu_{t-1}}
     \cdot \bm G_{\tau}(\bm{x}_{t},\bm{z}_\mathrm{sem},t)- \big(\bm{\tilde x}_0^{\mathrm target} - \bm{\hat x}_0^{\mathrm  student}(\bm x_t, t)\big) \rVert^2$ \COMMENT{Loss}
        \STATE ${\bm\theta} = {\bm\theta} - y \nabla_{{\bm\theta}}L_{{\bm\theta}}$ \COMMENT{Optimization}
    \ENDWHILE
\end{algorithmic}
\end{algorithm}

\begin{algorithm}[ht]
    \centering
    \caption{SFERD Sampling (with semantic gradient predictor)}\label{alg:sferd_sampling_dse}
    \begin{algorithmic}[0]
        \REQUIRE Trained student model $\bm S_\theta$, Trained Encoder $\mathrm{\bm E_\varphi}$
        , Trained Gradient Predictor $\bm G_{\tau}$
        \REQUIRE Sample $\bm x$
        
        \STATE$z_\mathrm{sem} = \mathrm{\bm E_\varphi}(\bm{x})$ 
        , $\bm{x}_{T} \sim \mathcal{N}(\bm{0}, \bm{\mathrm{I}})$ 
        \FOR{$t$ in $T, T-1, ..., 1$}
        \STATE$\bm{\tilde{\epsilon}}_{\theta}(\bm{x}_{t},t) = \bm{\epsilon}_{\theta}(\bm{x}_{t},t) - \sigma_{t} \cdot \bm{G}_{\tau}(\bm{x}_{t}, {z}_\mathrm{\bm sem}, t)$ 
        \STATE$\bm{x}_{t-1} = \mu_{t-1}\bigg(\frac{\bm{x}_{t} - \sigma_{t}{\bm{\tilde \epsilon}}_{\theta}(\bm{x}_{t},t)}{\mu_{t}} \bigg) + \sigma_{t-1} \cdot {\bm{\tilde \epsilon}}_{\theta}(\bm{x}_{t},t)$
        \ENDFOR
        \RETURN{$\bm{x}_{0}$} 
    \end{algorithmic}
\end{algorithm}

\section{Additional Implementation Details}
\lblapp{implementation-details}

\subsection{Experimental Details}
\lblapp{experimental-details}

The base diffusion model network we utilized is referenced from the NCSN++ architecture~\cite{song2020score} on CIFAR-10, and the BigGAN architecture~\cite{preechakul2022diffusion} on ImageNet 64$\times$64 and LSUN Bedroom
256$\times$256. We adopt identical parameterizations for SFERD-CD as those used for CD described in~\cite{song2023consistency} and SFERD-PD as those used for PD described in~\cite{salimans2022progressive}.
Table \ref{tab:pre-trained-PD}, \ref{tab:pre-trained-CD} show our detailed diffusion model network architecture and crucial parameterization settings of SFERD-PD and SFERD-CD.

\begin{table}[ht]
    \centering
    \begin{tabular}{l|cc}
    \toprule
    \textbf{Parameter} & \textbf{CIFAR 10} & \textbf{ImageNet 64x64}  \\
    \midrule
    Base channels & 64 & 128 \\
    Channel multipliers & {[}1,2,4,8{]} & {[}1,1,2,3,4{]}  \\
    Attention resolutions & \multicolumn{2}{c}{{[}16{]}} \ \ \ \ \\
    Attention heads num & 4 & \ 1  \\
    Batch size & 256 & \ 256  \\
    Learning rate & \multicolumn{2}{c}{2e-4 \ \ \ \ } \\
    Optimizer & \multicolumn{2}{c}{Adam (no weight decay)} \\
    Images trained & 64M & \ 96M  \\
    Diffusion loss & \multicolumn{2}{c}{\ \  MSE with noise prediction $\epsilon$} \\   
    \bottomrule
    \end{tabular}
    \vspace{2pt}
    \caption{Network architecture of pre-trained teacher models of SFERD-PD based on ADM~\cite{dhariwal2021diffusion}.}
    \label{tab:pre-trained-PD}
\end{table}

\begin{table*}[ht]
    \centering
    \begin{tabular}{l|ccc}
    \toprule
    \textbf{Parameter} & \textbf{CIFAR 10} & \textbf{ImageNet 64x64} & \textbf{Bedroom 256}  \\
    \midrule
    Base channels & 64 & 128 & 128 \\
    Channel multipliers & {[}1,2,4,8{]} & {[}1,1,2,3,4{]} & {[}1,1,2,3,4{]} \\
    Attention resolutions & \multicolumn{3}{c}{{[}16{]}} \ \ \ \ \\
    Attention heads num & 4 & \ 1 &  1 \\
    Batch size & 256 & \ 256 &  128 \\
   Learning rate & \multicolumn{3}{c}{2e-4 \ \ \ \ } \\
    Optimizer & \multicolumn{3}{c}{Rectified Adam~\cite{liuvariance} (no weight decay)} \\
     $\mu$ & 0 & 0.95 & 0.95 \\
     EMA decay rate & 0.9999 & 0.999943 & 0.999943 \\
    Images trained & 32M & \ 64M & 96M \\
    Diffusion loss & \multicolumn{3}{c}{\ \  MSE with noise prediction $\epsilon$} \\   
    \bottomrule
    \end{tabular}
    \vspace{2pt}
    \caption{Network architecture of pre-trained teacher models of SFERD-CD based on EDM~\cite{karras2022elucidating}. We use pre-trained DPMs provided by Consistency Distillation~\cite{song2023consistency}}
    \label{tab:pre-trained-CD}
\end{table*}

\begin{table*}[ht]
    \centering
    \begin{tabular}{l|cccc}
    \toprule
    \textbf{Parameter} & \textbf{CIFAR 10} & \textbf{ImageNet 64x64} & \textbf{Bedroom 256} \\
    \midrule
    MLP layers ($N$) & 10 & \ \ \ \ 10 & 20  \\
    MLP hidden size & \multicolumn{3}{c}{2048} \\
    Batch size & \multicolumn{3}{c}{512} \\
    Optimizer & \multicolumn{3}{c}{Adam (no weight decay)} \\
    Learning rate & \multicolumn{3}{c}{1e-4} \\
    EMA rate & \multicolumn{3}{c}{0.9999 / batch} \\
    $\beta$ scheduler & \multicolumn{3}{c}{Constant 0.008}  \\
    Diffusion loss & \multicolumn{3}{c}{L1 loss with noise prediction $\epsilon$} \\
    \bottomrule
    \end{tabular}
    \vspace{2pt}
    \caption{Network architecture of latent DPMs.}
    \label{latent}
\end{table*}

We use the Adam optimizer with a constant learning rate of $2 \times 10^{-4}$, no weight decay, and no dropout for SFERD-PD training on all datasets. 
We uniformly set the number of distillation stages to 2.
During the training of SFERD-CD, we also disabled weight decay, learning rate warmup, dropout, data augmentation, and gradient clipping for all settings. Compared to DDIM, EDM requires lower NFEs. We distill the teacher model from 40 steps (79 NFEs with Runge-Kutta) for CIFAR-10, and from 128 timesteps (255 NFEs with Runge-Kutta) for ImageNet 64$\times$64 and Bedroom 256$\times$256 to train one-step or multi-step student models. 
All models are pre-trained in-house, except for CD and EDM on ImageNet 64$\times$64 and Bedroom 256$\times$256, which use the provided checkpoint from~\cite{song2023consistency}. 
All experiments are conducted on 8 NVIDIA GeForce RTX 3090 GPUs. Moreover, We apply horizontal flips for data augmentation strategy to all models and datasets. 

Regarding the design of the student model with semantic gradient predictor, we use the encoder part of U-Net~\cite{ronneberger2015u} for encoder $\bm{E}_{\varphi}$.
Considering that $\bm{\tilde x}_0^{\mathrm target}$ also takes $\bm{x}_{t}$ and $t$ as input, we utilize the teacher's knowledge by reusing its trained encoder part and time embedding layer. This allows us to only require new middle, decoder, and output blocks of U-Net to construct $\bm{G}_{\tau}$.
To incorporate $\bm{z}$ into them, we follow~\cite{dhariwal2021diffusion} to extend Group Normalization~\cite{wu2018group} by applying scaling \& shifting twice on normalized feature maps:
\begin{equation}
    \begin{aligned}
        \text{AdaGN}(\bm{h},t,\bm{z}) = \bm{z}_{s} (t_{s} \text{GroupNorm}(\bm{h}) + t_{b}) + \bm{z}_{b} \,,
    \end{aligned}
\end{equation}
where $[\bm{t}_{s}, \bm{t}_{b}]$ and $[\bm{z}_{s}, \bm{z}_{b}]$ are obtained from a linear projection of $t$ and $\bm{z}$, respectively. We still utilize the skip connections from the reused encoder to the new decoder. As a result, $\bm{G}_{\tau}$ is determined exclusively by the pre-trained teacher model, which allows for its application to different U-Net architectures.

We use deep MLPs as the denoising network of latent DPMs.
Table \ref{latent} shows the network architecture.
Specifically, we calculate $\bm{z}_{\mathrm{sem}}=\bm{E}_{\varphi}(\bm{x}_{0})$ for all $\bm{x}_{0}$ from datasets and normalize them to zero mean and unit variance.
Then we learn the latent DPMs $p_{\omega}(\bm{z}_{t-1} | \bm{z}_{t})$ by optimizing:
\begin{equation}
    \begin{aligned}
        \mathcal{L}(\kappa) = \mathbb{E}_{\bm{z},t,\epsilon}\big[ \| \epsilon - \bm{\epsilon}_{\omega}(\bm{z}_{sem}^{t},t) \| \big] \,,
    \end{aligned}
\end{equation}
where $\epsilon\sim \mathcal{N}(\bm{0}, \bm{\mathrm{I}})$ and $\bm{z}_mathrm{sem}^{t} = \sqrt{\bar{\alpha}_{t}}\bm{z}_\mathrm{sem}+\sqrt{1-\bar{\alpha}_{t}}\epsilon$.
The sampled $\bm{z}$ will be denormalized for use.

\begin{figure}[h]
    \centering
    \includegraphics[width = 1.01\columnwidth]{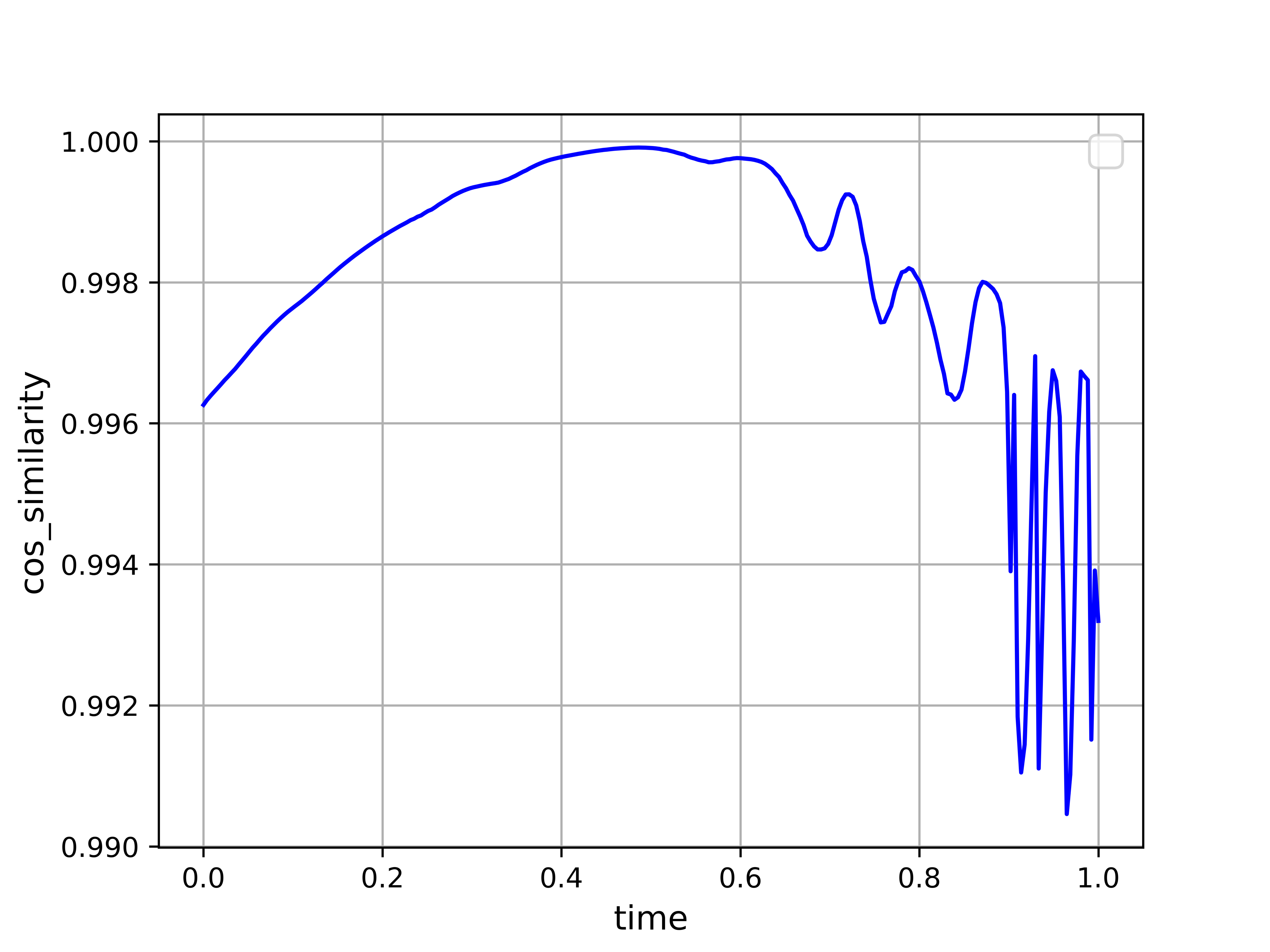}
    \caption{The cosine similarity between $\nabla_{\bm{x}_{t}} \log p_{\tau}(\mathrm{E_\varphi}(\bm x_0)|\bm{x}_t)$ and $G_\tau(\bm x_t,\mathrm{E_\varphi}(\bm x_0),t)$ and at different times $t$ in pre-trained student model with semantic gradient predictor.}
    \label{fit_gradient}
\end{figure}

 \begin{table}[ht]
\centering
\begin{tabular}{lrr}
\noalign{\smallskip}\noalign{\smallskip}\toprule
{Model} & {NFE} & FID ($\downarrow$)\\
\midrule
  SFERD-CD (ours)& 1 &9.38  \\ 
 CD~\cite{song2023consistency} & 1 & 16.09 \\
StyleGAN2~\cite{Karras2019AnalyzingAI} & 1 & \textbf{2.35} \\
PG-SWGAN~\cite{Wu2019SlicedWG} & 1 & 8.00\\
 \midrule
 SFERD-CD (ours) & 2 & \textbf{5.62} \\
 CD~\cite{song2023consistency} & 2 & 6.37  \\
  \midrule
 SFERD-CD (ours) & 4 & \textbf{4.92} \\

 CD~\cite{song2023consistency} & 4 & 5.59  \\
 

\bottomrule
\end{tabular}
\vspace{4pt}
\caption{Sample quality on Bedroom 256$\times$256. SFERD-* represents the implementation of the corresponding model within SFERD framework. Both attention guidance method and semantic encoding-based gradient predictor are introduced.}
\label{tab:bed}
\end{table}

\begin{figure*}[ht]
\label{DSE_derivation}
\begin{align}
    \notag \mathcal{L}(\theta, \varphi, \tau) &=\mathbb E_{\bm{x}_{0},\epsilon}\bigg[\sum_{t>1} D_{\mathrm KL}\big(q_{t}(\bm x_{t-1}|\bm x_t,\bm {\tilde x}_0^{target})||p_{\theta}(\bm x_{t-1}|\bm x_t, \bm z_{sem})\big)+C\bigg]
    \\\notag &= 
    \mathbb{E}_{\bm{x}_{0},t,\epsilon}\bigg[\omega(\lambda_t)\big\| \bm{\Sigma}_{\theta}(\bm{x}_{t}, t) \cdot \nabla_{\bm{x}_{t}} \log p_{\tau}(\bm{z}_{sem}|\bm{x}_t) - \big(\widetilde{\bm{\mu}}_{t}(\bm{x}_{t},\bm{\tilde x}_{0}^{target}) - \bm{\mu}_{\theta}(\bm{x}_{t},t)\big)  \big\|^{2}\bigg] 
    \\\notag &= 
    \mathbb{E}_{\bm{x}_{0},t,\epsilon}\bigg[\omega(\lambda_t)\big\| \bm{\Sigma}_{\theta}(\bm{x}_{t}, t) \cdot  \nabla_{\bm{x}_{t}} \log p_{\tau}(\bm{z}_{sem}|\bm{x}_t)  - \frac{\sqrt{\bar \alpha_{t-1}}\beta_t}{1-\bar \alpha_t}
    \big(\bm{\tilde x}_0^{\mathrm target} - \bm{\hat x}_0^{\mathrm  student}(\bm{x}_{t},t)\big)   \big\|^{2}\bigg]
    \\\notag &=
    \mathbb{E}_{\bm{x}_{0},t,\epsilon}\bigg[\omega(\lambda_t)\big\| \bm{\Sigma}_{\theta}(\bm{x}_{t}, t) \cdot \big(\nabla_{\bm{x}_{t}} \log p_{\tau}(\bm{x}_t | \bm{z}_{sem}) - 
    \nabla_{\bm{x}_{t}} \log p_{\tau}(\bm{x}_t)\big)\\\notag & - \frac{\sqrt{\bar \alpha_{t-1}}\beta_t}{1-\bar \alpha_t}
    \big(\bm{\tilde x}_0^{\mathrm target} - \bm{\hat x}_0^{\mathrm  student}(\bm{x}_{t},t)\big) \big\|^{2}\bigg]
    \\\notag &\approx
    \mathbb{E}_{\bm{x}_{0},t,\epsilon}\bigg[\omega(\lambda_t)\big\| -\frac {\bm{\Sigma}_{\theta}(\bm{x}_{t}, t)}{\sqrt{1-\bar \alpha_t}} \cdot \big(-\sqrt{1-\bar \alpha_t}\nabla_{\bm{x}_{t}} \log p_{\tau}(\bm{x}_t | \bm{z}_{sem}) \\\notag & + \sqrt{1-\bar \alpha_t}
    \nabla_{\bm{x}_{t}} \log p_{\tau}(\bm{x}_t)\big) -
    \big(\bm{\tilde x}_0^{\mathrm target} - \bm{\hat x}_0^{\mathrm  student}(\bm{x}_{t},t)\big) \big\|^{2}\bigg]
    \\\notag &\approx
    \mathbb{E}_{\bm{x}_{0},t,\epsilon}\bigg[\omega(\lambda_t)\big\| -\frac {\bm{\Sigma}_{\theta}(\bm{x}_{t}, t)}{\sqrt{1-\bar \alpha_t}} \cdot \frac{1-\bar \alpha_t}{\sqrt{\bar \alpha_{t-1}}\beta_t} \cdot \big(-\sqrt{1-\bar \alpha_t}\nabla_{\bm{x}_{t}} \log p_{\tau}(\bm{x}_t | \bm{z}_{sem}) \\\notag &+  \sqrt{1-\bar \alpha_t}
    \nabla_{\bm{x}_{t}} \log p_{\tau}(\bm{x}_t)\big) - 
    \big(\bm{\tilde x}_0^{\mathrm target} - \bm{\hat x}_0^{\mathrm  student}(\bm{x}_{t},t)\big) \big\|^{2}\bigg]
    \\\notag &\approx
    \mathbb{E}_{\bm{x}_{0},t,\epsilon}\bigg[\omega(\lambda_t)\big\| -\frac {\sqrt{1-\bar \alpha_t}\bm{\Sigma}_{\theta}(\bm{x}_{t}, t)} {\sqrt{\bar \alpha_{t-1}}\beta_t} \cdot \big(\epsilon_\theta(\bm{x}_t , \bm{z}_{sem},t) -  \epsilon_\theta(\bm{x}_t, t)\big) - 
    \big(\bm{\tilde x}_0^{\mathrm target} \\\notag & - \bm{\hat x}_0^{\mathrm  student}(\bm{x}_{t},t)\big) \big\|^{2}\bigg]
    \\\notag&\approx
    \mathbb{E}_{\bm{x}_{0},t,\epsilon}\bigg[\omega(\lambda_t)\big\| -\frac {\sqrt{1-\bar \alpha_t}\bm{\Sigma}_{\theta}(\bm{x}_{t}, t)} {\sqrt{\bar \alpha_{t-1}}\beta_t} \cdot \bigg(\frac{\bm x_t - \sqrt{\bar \alpha_t}\bm{\hat x}_0^{\mathrm  student}(\bm{x}_{t},\bm{z}_{sem},t)}{\sqrt{1-\bar\alpha_t}} \\\notag &  - \frac{\bm x_t - \sqrt{\bar \alpha_t}\bm{\hat x}_0^{\mathrm  student}(\bm{x}_{t},t)}{\sqrt{1-\bar\alpha_t}}\bigg)  - \big(\bm{\tilde x}_0^{\mathrm target} - \bm{\hat x}_0^{\mathrm  student}(\bm{x}_{t},t)\big)
    \big\|^{2}\bigg]
    \\&\notag \approx
    \mathbb{E}_{\bm{x}_{0},t,\epsilon}\bigg[\omega(\lambda_t)\big\| \frac{(1-\bar \alpha_{t-1})\sqrt{\bar \alpha_t}}{\bar \alpha_{t}}
     \cdot \bigg( \bm{\hat x}_0^{\mathrm  student}(\bm{x}_{t},\bm{z}_{sem},t) - \bm{\hat x}_0^{\mathrm  student}(\bm{x}_{t},t) \bigg) \\ &- \big(\bm{\tilde x}_0^{\mathrm target} - \bm{\hat x}_0^{\mathrm  student}(\bm{x}_{t},t)\big)\|^{2}\bigg]
\end{align}
\end{figure*}

\subsection{Distillation with semantic gradient predictor}
\lblapp{RD-z}

We provide the detailed derivation of the training loss based on the student model with semantic gradient predictor as Eq.(18). According to Bayes' rule, we have $\log p_{\tau}(\bm{z}_{\mathrm {sem}}|\bm{x}_t) \propto \log \big( p_{\tau}(\bm{x}_t|\bm{z}_{\mathrm {sem}}) / p_{\tau}(\bm{x}_t)\big)$. Thus, we can further express
$\nabla_{\bm{x}_{t}} \log p_{\tau}(\bm{z}_{\mathrm {sem}}|\bm{x}_t) =  \nabla_{\bm{x}_{t}} \log p_{\tau}(\bm{x}_t | \bm{z}_{\mathrm {sem}}) -\nabla_{\bm{x}_{t}} \log p_{\tau}(\bm{x}_t)$.

We exhibit the effectiveness of the semantic gradient predictor in the experiment section. In order to confirm whether $G_\tau$ accurately fits $\nabla_{\bm{x}_{t}} \log p_{\tau}(\bm{z}_\mathrm{sem}|\bm{x}_t)$,
we randomly select 500 real images $\bm x_0$ from ImageNet 64$\times$64 and obtain their corresponding semantic representations $\bm z_{sem} = E_\varphi(\bm x_0)$. We compute the corresponding gradients $\nabla_{\bm{x}_{0}} \log p_{\tau}(\bm{z}_\mathrm{sem}|\bm{x}_0)$. Next, we apply the forward diffusion process using the equation $\bm x_t=\mu_t \bm x_0+\sigma_t \bm \epsilon$ for a total of 1024 steps (T=1024). Further based on Eq.(\ref{predictor}), we calculate the cosine similarity between $\nabla_{\bm{x}_{t}} \log p_{\tau}(\bm{z}_\mathrm{sem}=E_\varphi(\bm x_0)|\bm{x}_t)$ and the predicted gradient of $G_\tau$. 
\begin{equation}
\label{predictor}
\begin{aligned}
    \nabla_{\bm{x}_{t}} \log p_{\tau}(\mathrm{E_\varphi}(\bm x_0)|\bm{x}_t) & := G_\tau(\bm x_t,\mathrm{E_\varphi}(\bm x_0),t) \\ & = \nabla_{\bm{x}_{0}} \log p_{\tau}(\mathrm{E_\varphi}(\bm x_0)|\bm{x}_0) \cdot \frac{\partial \bm x_0}{\partial \bm x_t}
\end{aligned}
\end{equation}
As shown in Figure~\ref{fit_gradient}, it can be observed that the cosine similarity between the two variables remains consistently above 0.99 throughout the process, especially during the middle stage, the fitting shows the best performance. It suggests that $G_\tau$ is a good fit for $\nabla_{\bm{x}_{t}} \log p_{\tau}(\bm{z}_{sem}|\bm{x}_t)$ despite not being the direct regression target during the process. 
Furthermore, to enhance the training convergence speed, we can carefully and moderately reduce the scale of $G_\tau$. However, this would lead to a decline in the quality of generation. It is crucial to achieve a balance.
The results of our experiment further confirm that decreasing the scale of $\mathrm{E_\varphi}$ does not result in a significant decline in quality.
In fact, the idea behind the student model with semantic gradient predictor is accidentally similar to ControlNet (Zhang 2023). We opt to use $z_\mathrm{sem}$ as an input for an additional prediction network and jointly train it with the frozen previous trained distillation student model. Compared to directly using it to optimize the original weights of the student model, this choice would help to avoid overfitting.

\begin{figure}[ht]
    \centering
\includegraphics[width=1.01\linewidth]{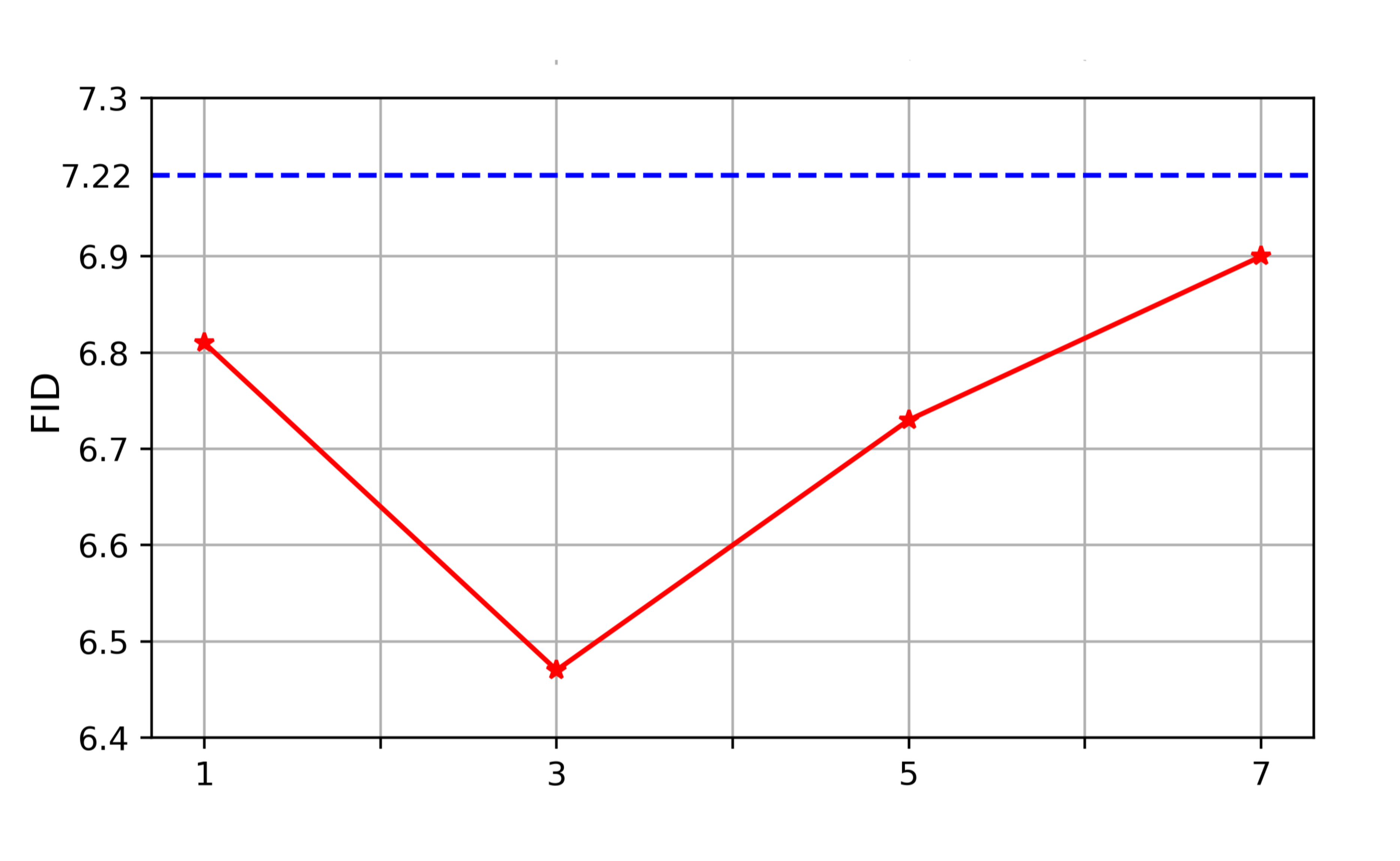}
        \caption{\textbf{Sigma ($\sigma$) of Gaussian blur.}}
        \label{blur_sigma}
\end{figure}

\subsection{Additional Experiments}
\lblapp{add-experiment}
\subsubsection{Distillation stages.} Due to the fitting error of the teacher during the sampling process, there exists a gap between the predicted noise $\bm \epsilon_\theta(\bm{\hat x}_t, t)$ and the true noise $\bm \epsilon$, even when employing accurate numerical solvers. 
This difference accumulates gradually and is introduced in subsequent iterations. The cumulative error is further exacerbated by the multiple distillation training stages for the diffusion distillation model, affecting the student's performance. However, using too large a distillation step can also easily lead to the disappearance of gradients, making the training loss difficult to reduce. 
Thus a good choice of distillation step should be a trade-off. 
We present an ablation experiment, varying the final sampling step and the number of distillation stages in SFERD-PD utilizing the initial teacher step of 1024 trained on the CIFAR-10. 
We refer to the default training setting of the original PD, using a training image length of 32M for each distillation stage. The results are presented in Table \ref{tab:distill_stage}. It can be observed that the best performance is achieved for the two-stage distillation, regardless of the number of sampling steps. Neither too large nor too small distillation stages could improve the performance.
\begin{table}[h]
\renewcommand{\arraystretch}{0.9}
\setlength{\tabcolsep}{3pt} 
\centering
\small
\begin{tabular}{lcrr}
\noalign{\smallskip}\noalign{\smallskip}\toprule
Schedule & Stage Number & Training Length & FID($\downarrow$) \\
\midrule
 1024, 1 &   1  & 32M & 16.79  \\
 1024, 32, 1 &  2  & 64M & \textbf{7.54} \\
 1024, 256, 32, 8, 1  & 4 & 128M &  9.31 \\
 1024, 2 &  1  & 32M & 14.22 \\
 1024, 64, 2  &  2   & 64M &  \textbf{6.37} \\
 1024, 256, 64, 8, 2  & 4 & 128M & 8.28\\
\bottomrule
\end{tabular}
\caption{\textbf{Ablation study of the number of distillation stage.}} \vspace{-10pt}
\label{tab:distill_stage}
\end{table}

\subsubsection{Implementation on Bedroom 256.}Due to the limit of computational resources, we only implement SFERD-CD on LSUN Bedroom 256$\times$256. The result is shown in Table~\ref{tab:bed}, which indicates that SFERD-CD achieves better performance than the original model at various steps.
Additionally, we provide error bars for the evaluation metrics computed in the experiment section of the main paper (Table~\ref{tab:i64_cifar10}).

\section{Exploration in Diffusion Models}
\lblapp{interior-exploration}
\subsection{Predicted Noise Analysis}
\begin{figure*}[h]
    \centering
    \includegraphics[width = .72\textwidth]{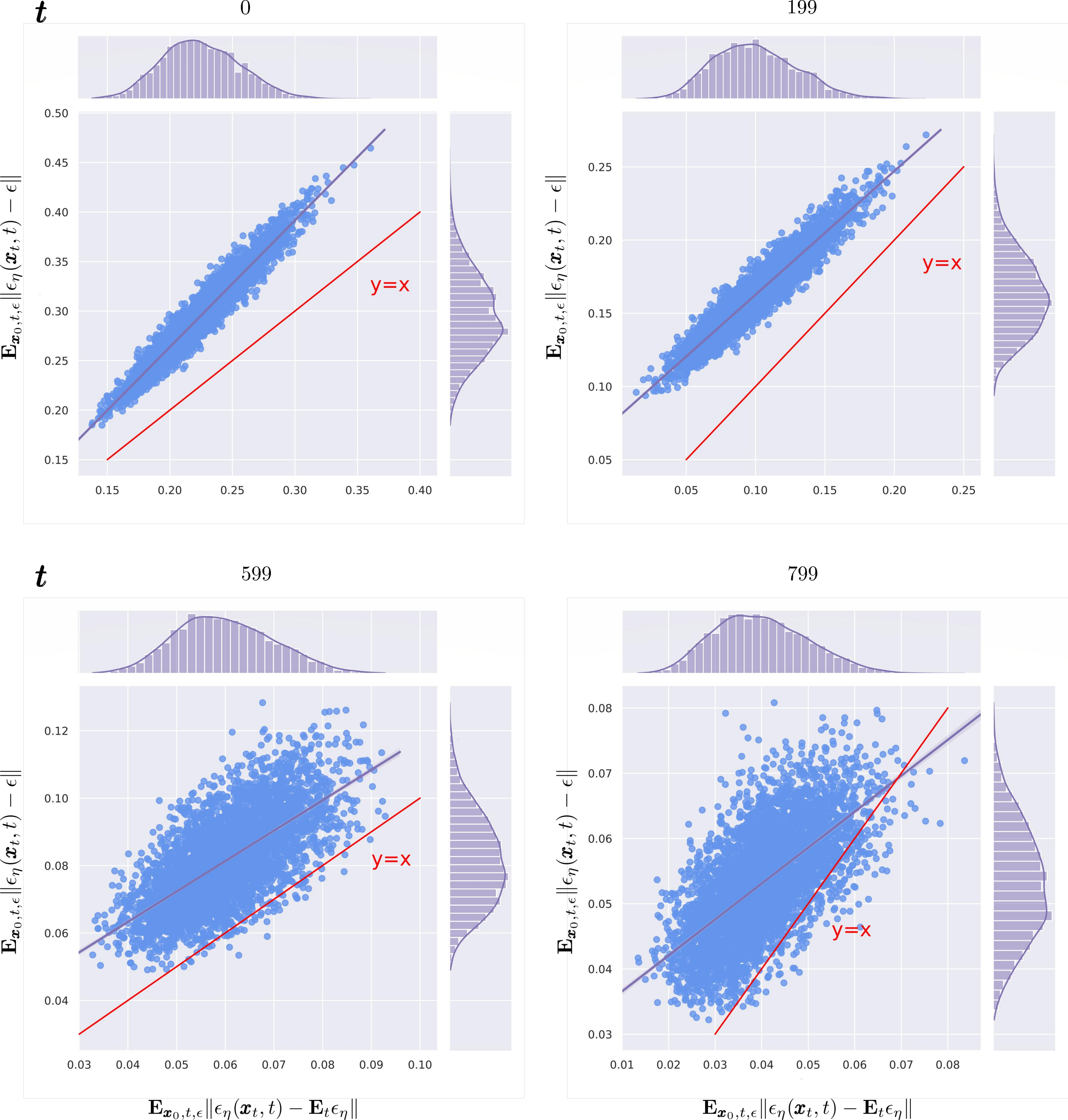}
    \caption{The correlation trend between the values of $\mathbb{E}\| \epsilon_\eta(\bm x_t, t) - \epsilon \|$ (fitting error) and $\mathbb{E} \| \epsilon_\eta(\bm x_t, t)  - \mathbb{E}_t \epsilon_\eta(\bm x_t, t) \|$ (variance) at specify time $t$ using $\epsilon$-prediction pre-trained diffusion model on ImageNet 128$\times$128, given $\bm x_0$ and $\bm \epsilon$.}
    \label{app_fluctation}
\end{figure*}

\lblapp{analysis-eps}
Figure~\ref{app_fluctation} shows the rest correlation trend and distribution between the values of $\mathbb{E}\| \epsilon_\eta(\bm x_t, t) - \epsilon \|$ (fitting error) and $\mathbb{E} \| \epsilon_\eta(\bm x_t, t)  - \mathbb{E}_t \epsilon_\eta(\bm x_t, t) \|$ (variance) at specify time $t$. Table~\ref{tab:error_ms} displays the means and standard deviations of the fitting error and variance at different $t$. It is observed that the fitting error is primarily determined by the variance, which spans the entire global range. Furthermore, the fitting error displays a positive correlation with the variance.

\begin{figure}[h]
    \centering
    \begin{minipage}[t]{0.51\textwidth}
        \centering
    \includegraphics[width = .95\textwidth]{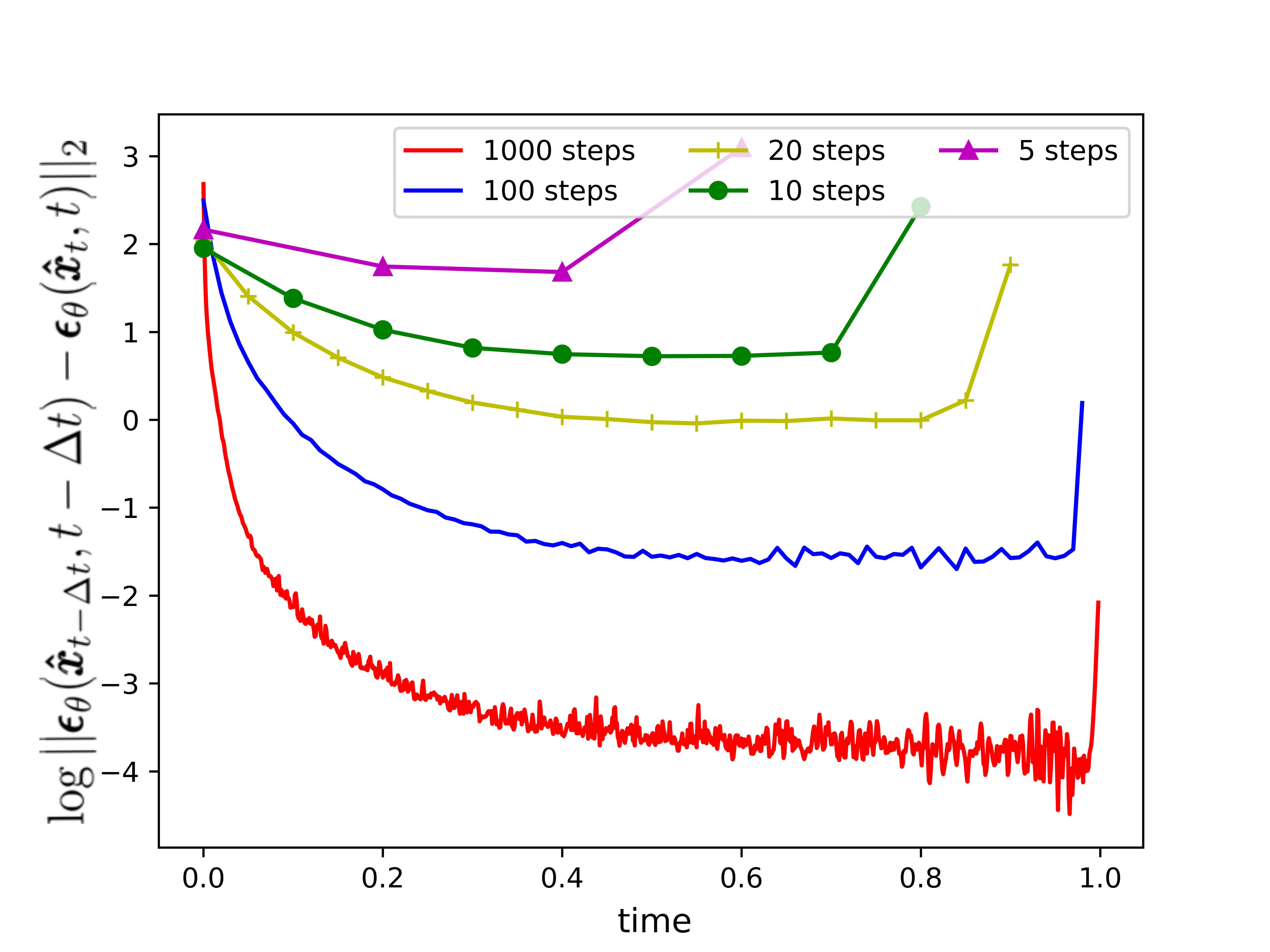}
    \caption{Distance between neighboring predicted noises at different sampling timesteps on the CIFAR-10 dataset, using the $\bm \epsilon$-prediction pre-trained diffusion model.}
    \label{delta_eps}
    \end{minipage}
    \hspace{.05in}
    \begin{minipage}[t]{0.40\textwidth}
        \centering
        \includegraphics[width=0.95\textwidth]{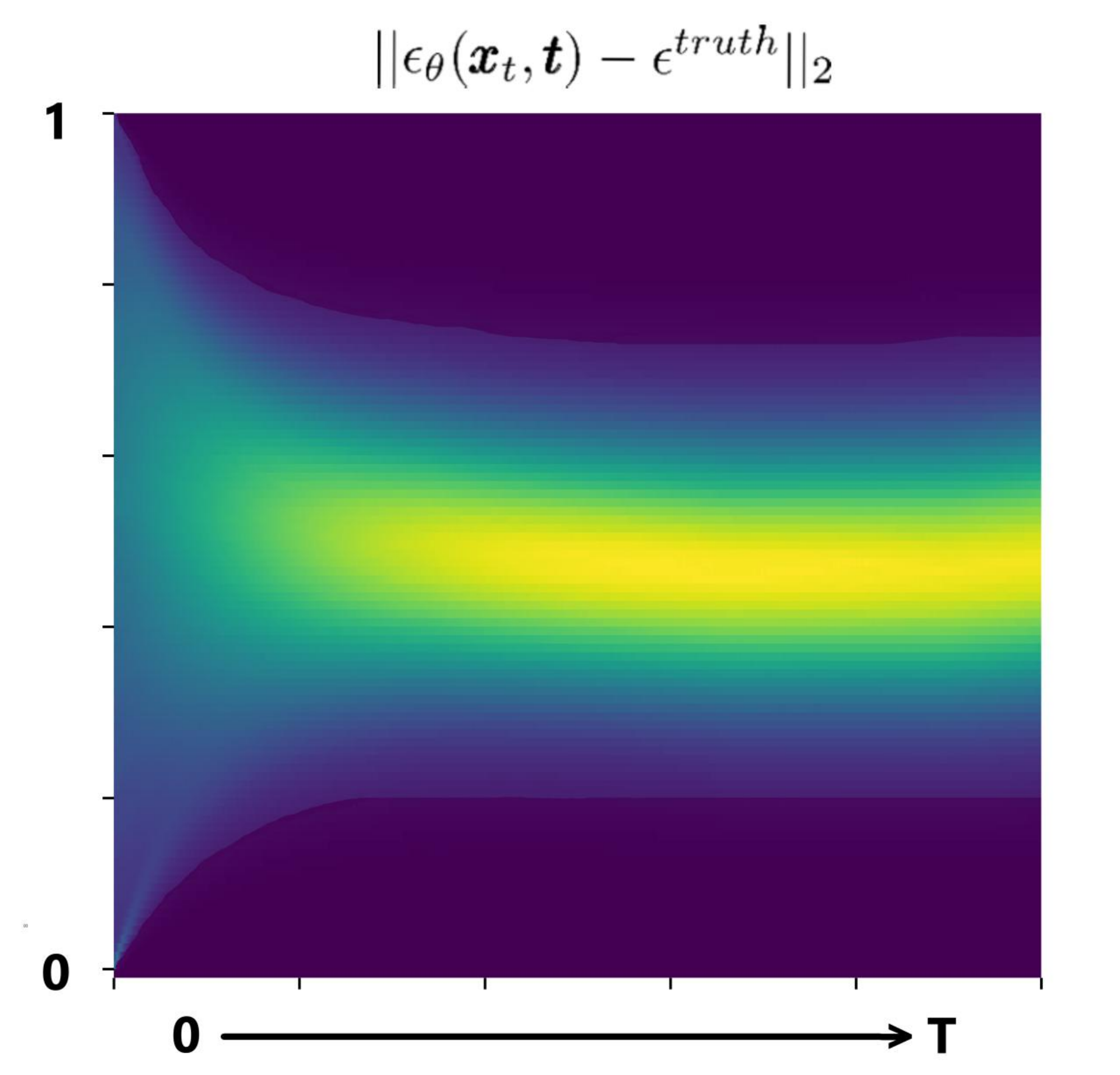}
        \caption{Fitting error during $\epsilon$-prediction diffusion model training on CIFAR-10. Lighter areas represent the smaller fitting error.}
        \label{heat_eps}
    \end{minipage}
\end{figure}

Figure~\ref{delta_eps} shows the L2 distance between adjacent predicted noises of the $\bm{\epsilon}$-prediction pre-trained diffusion model trained on CIFAR-10 using DDIM at different total sampling steps. We observe that as the total sampling steps decrease, the adjacent predicted noise fluctuates more violently, which is aligned with the analysis in Eq (\ref{ode-4}). Thus, maintaining a smooth change in predicted noise during large stepsize inference is critical to guarantee image generation quality. 
The distribution of real data is typically sparse, so the prediction network may perform poorly due to insufficient local training data, especially in the areas where $p_t(\bm x)$ is small, resulting in the high fitting error (Figure~\ref{heat_eps}). Therefore, To ensure that $\bm{\hat x}_0$ is close to real image $\bm x_0$, it is also essential to keep $\bm x_t$ in the high $p_t(\bm x)$ regions for all $t$.

\begin{table}[ht]
\centering
\begin{tabular}{lrrrr}
\noalign{\smallskip}\noalign{\smallskip}\toprule
\multirow{2}{*}{t} &\multicolumn{2}{c}{The fitting error} &\multicolumn{2}{c}{The variance}\\
  & Mean  & Std & Mean & Std  \\
  \midrule
  0 & 0.2956  & 0.0413  & 0.2250 & 0.0313 \\ 
  199 & 0.1627 & 0.0266 & 0.1048 & 0.0182\\ 
 399  &0.1121 & 0.0187 & 0.0754 & 0.0119 \\
 599 & 0.0813 & 0.0133 & 0.0569 & 0.0094 \\
 799  & 0.0531 & 0.0084 & 0.0485 & 0.0076\\

\bottomrule
\end{tabular}
\vspace{3pt}
\caption{The means and standard deviations of $\mathbb{E}\| \epsilon_\eta(\bm x_t, t) - \epsilon \|$ (fitting error) and $\mathbb{E} \| \epsilon_\eta(\bm x_t, t)  - \mathbb{E}_t \epsilon_\eta(\bm x_t, t) \|$ (variance) at specify time $t$ using $\epsilon$-prediction pre-trained diffusion model on ImageNet 128$\times$128.}
\label{tab:error_ms}
\end{table}

\subsection{Attention Analysis}
\lblapp{analysis-attention}

The method for upsampling the attention map within the diffusion model is presented as Eq.(\ref{Attn_map}). First, global average pooling (GAP) is performed, then reshape into the input image $\bm{\hat x}_t$ dimensions (CxHxW), and finally use the nearest neighbor upsampling method to match the resolution of $\bm{\hat x}_t$. Although attention maps cannot capture the intact information from images, they do discover and highlight the general semantic parts or regions at the corresponding time.
\begin{equation}\label{Attn_map}
\begin{aligned}
    \mathrm {Amap}_t^l = \rm{Upsample}\big(\mathrm {Reshape}(GAP(A_t^{l})\big)
\end{aligned}
\end{equation}

\subsection{Validation in Reducing the Fitting Error}
To evaluate the effectiveness of our proposed methods in reducing the fitting error of the diffusion distillation model, we randomly select some real images from CIFAR-10 and ImageNet 64$\times$64 and perform forward diffusion to obtain $\bm x_t$. We then calculate the means and standard deviations of the fitting error $\mathbb{E}_{\bm x_0,t,\epsilon}\| \epsilon(\bm x_t, t) - \epsilon \|$ for the teacher and student models of PD and SFERD-PD pre-trained on CIFAR10 and ImageNet 64$\times$64, respectively. The results are shown in Table~\ref{tab:validation_t} and Table~\ref{tab:validation_s}. All results are averaged over 500 random images and the entire period. The total diffusion stepsize for the teacher model is 1024, and for the student model, it is 4 steps. It is evident that the introduction of attention guidance and semantic gradient predictor leads to the reduction of the fitting error in both teacher and student models, compared to the original distillation model. 
In particular, the enhancement in the student model is more pronounced.

\begin{table*}[h]
\centering
    \begin{tabular}{lrrrr}
\noalign{\smallskip}\noalign{\smallskip}\toprule
    \multirow{2}{*}{Model} &\multicolumn{2}{c}{CIFAR-10}&\multicolumn{2}{c}{ImageNet64} \\
      & Mean ($\downarrow$) & Std ($\downarrow$)  &  Mean ($\downarrow$) & Std ($\downarrow$)\\
     \midrule
      teacher of PD   &0.00896 & 0.01290 & 0.0480 & 0.0783 \\ 
     teacher of SFERD-PD   & \textbf{0.00863} &\textbf{0.01277} & \textbf{0.0426} & \textbf{0.0711} \\ 
     
    \bottomrule
    \end{tabular}
    \vspace{3pt}
    \caption{The means and standard deviations of the fitting error $\mathbb{E}_{\bm x_0,t,\epsilon}\| \epsilon_\eta(\bm x_t, t) - \epsilon \|$  for the teacher models of PD and SFERD-PD pre-trained on CIFAR10 and ImageNet 64$\times$64, respectively.}
    \label{tab:validation_t}
\end{table*}

\begin{table*}[h]
\centering
\begin{tabular}{lrrrr}
\noalign{\smallskip}\noalign{\smallskip}\toprule
\multirow{2}{*}{Model} &\multicolumn{2}{c}{CIFAR-10}&\multicolumn{2}{c}{ImageNet64} \\
  & Mean ($\downarrow$)  & Std ($\downarrow$) &  Mean ($\downarrow$)  & Std ($\downarrow$)\\
 \midrule
    student of PD  & 0.01147 & 0.01896 & 0.0523 & 0.1082 \\ 
  student of SFERD-PD  &\textbf{0.00908} &\textbf{0.01329} & \textbf{0.0440} & \textbf{0.0768}\\ 
\bottomrule
\end{tabular}
\vspace{3pt}
\caption{The means and standard deviations of the fitting error $\mathbb{E}_{\bm x_0,t,\epsilon}\| \epsilon_\theta(\bm x_t, t) - \epsilon \|$ for the studnet models of PD and SFERD-PD pre-trained on CIFAR10 and ImageNet 64$\times$64, respectively.}
\label{tab:validation_s}
\end{table*}

     
\begin{figure}[ht]
    \centering
    \includegraphics[width = .99\columnwidth]{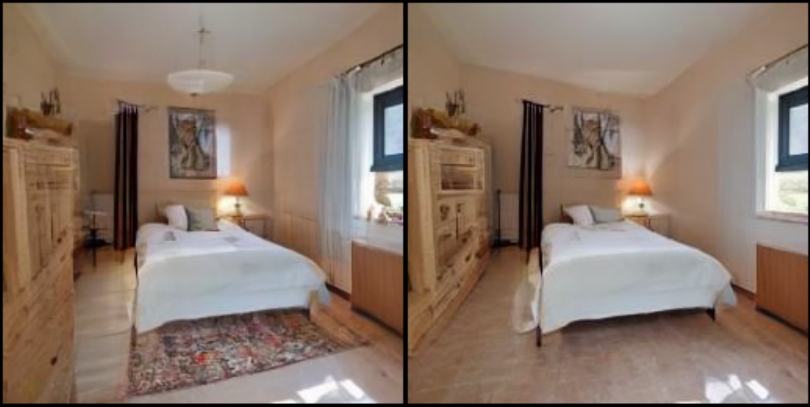}
    \caption{An example pair from our evaluation samples. The participants are asked \textbf{which side do you think shows the better quality (left or right ?)}, without being informed which side is sampled with our method (random).}
    \label{human_eval}
\end{figure}

\section{Human Evaluation}
\lblapp{human-evaluation}
For the human evaluation of SFERD with samples from Bedroom 256$\times$256, we generate a total of 1000 pairs using SFERD-CD and CD with 4 sampling steps. Each pair shares the same seed of the initial noise for a fair comparison. We presented each of the 1000 participants with a pair of two groups, one sample in each group, with one group using SFERD-CD and the other using CD. Participants are then asked to evaluate and select a group with higher image quality. An example is shown in Figure~\ref{human_eval}. Each pair of samples is evaluated five times. The samples are neither cherry-picked nor filtered. Additionally, we do not post-process the responses in any way. The results indicate that 52.8 ($\pm 1.1$)\% of the participants perceive the images generated by SFERD-CD model to have higher quality.


\begin{figure}[t]
    \centering
    \begin{minipage}[h]{0.49\textwidth}
        \includegraphics[width=0.95\textwidth]{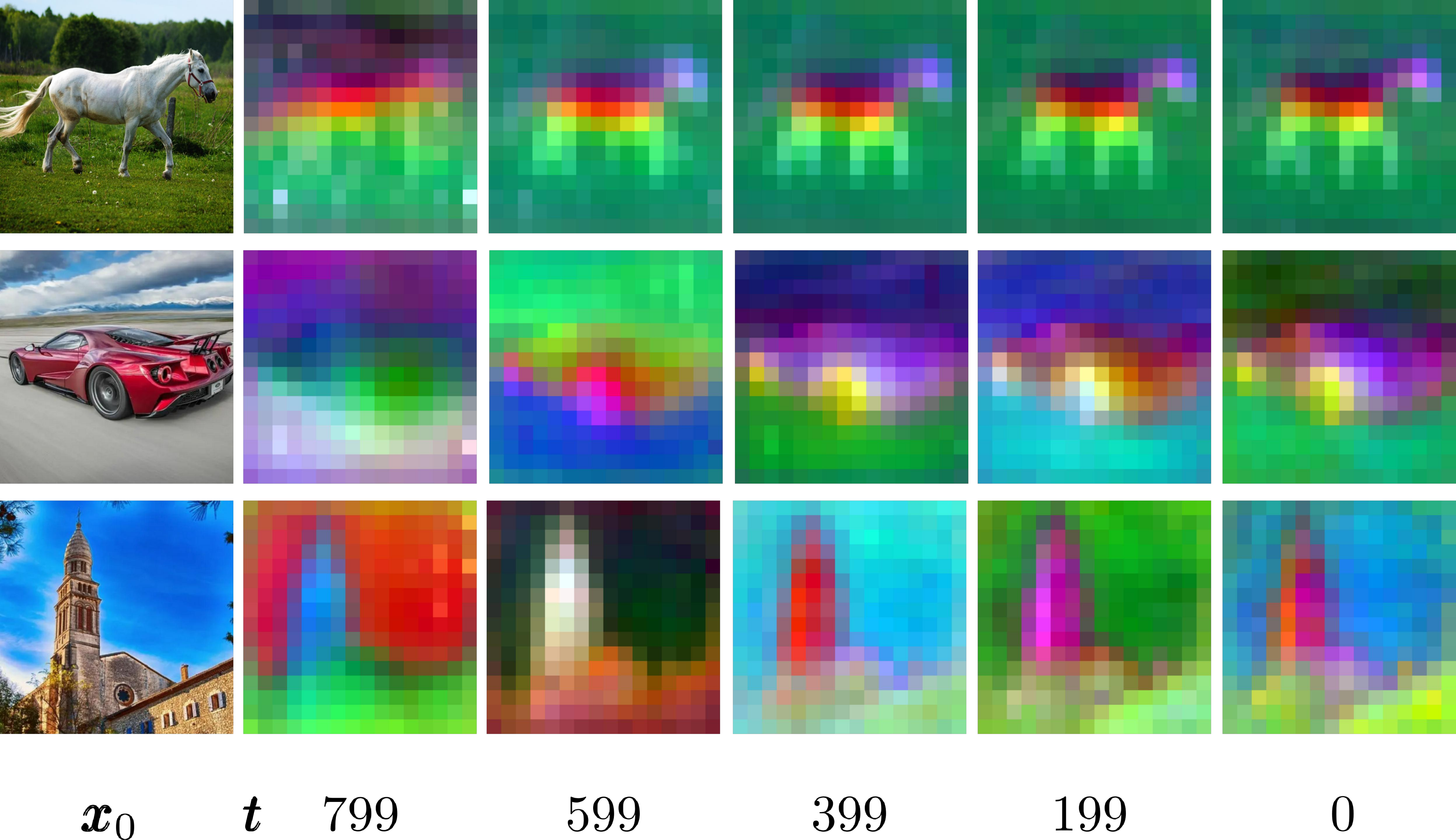}
        \caption{
        Visualising diffusion features of $\epsilon_\theta(\bm {x}_t, t)$ from decoder (layers 4) in $\epsilon$-prediction pre-trained diffusion model on ImageNet 128$\times$128. We applied PCA on the extracted features across all images and visualized the top three leading components. 
        }
        \label{delta_eps_l1}
    \end{minipage}%
    \hspace{.03in}
    \begin{minipage}[h]{0.49\textwidth}
        \includegraphics[width = .95\textwidth]{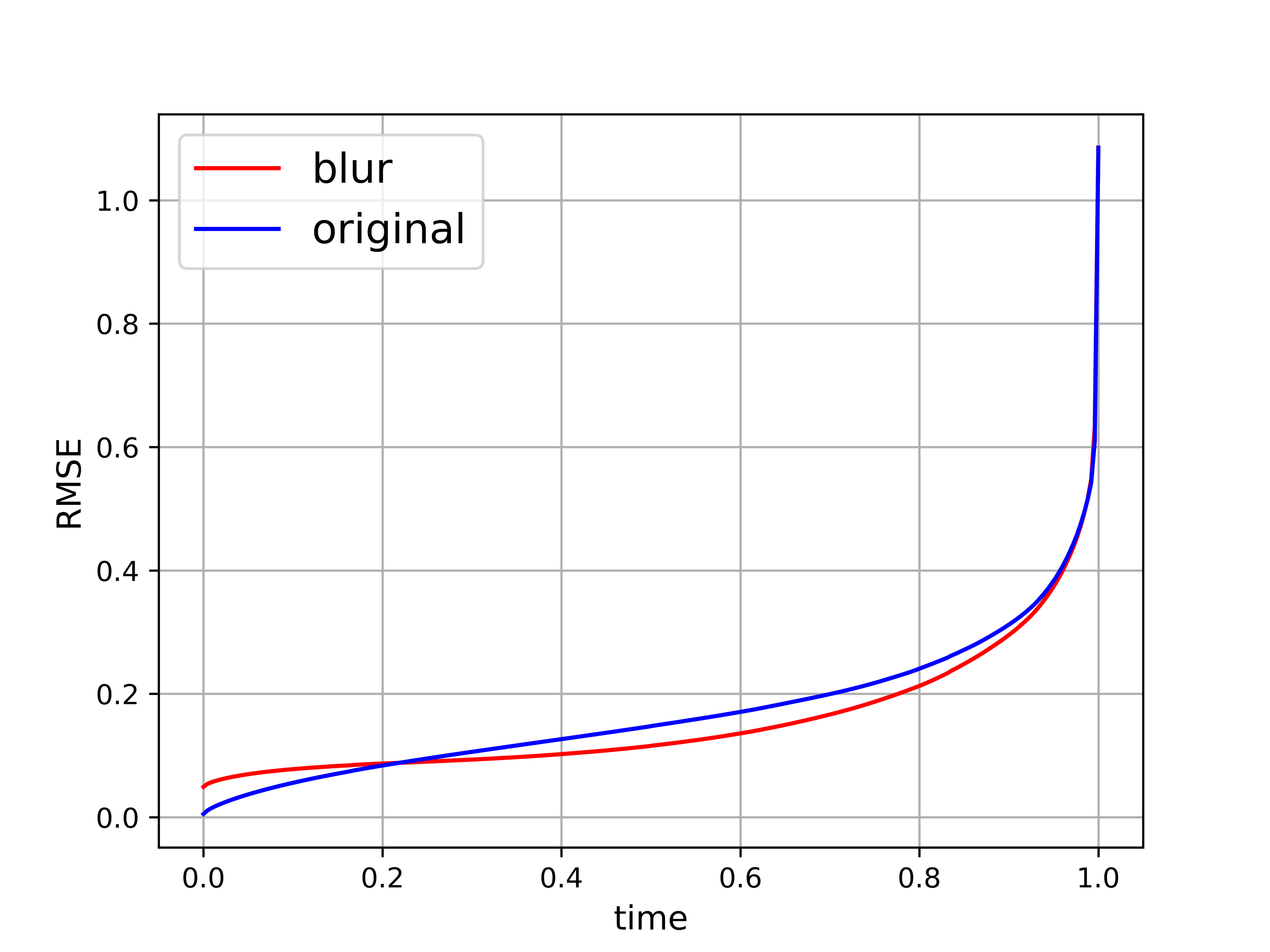}
        \caption{The comparison of RMSE between $\textrm{Gaussian\text-Blur}(\bm{\hat{x}}_0^{t})$ (blur strength $\sigma=3$) and $\bm{\hat{x}}_0^{t}$ (original) to the final generated images using $\epsilon$-prediction pre-trained teacher model on ImageNet 64 $\times$ 64.}
        \label{blur_x0}
    \end{minipage}%
\end{figure}
\section{Discussion $\&$ Limitation}
\lblapp{limitation}
\subsection{Gaussian Blur Discussion}
We discuss the reasons for using Gaussian blur by point.
\begin{itemize}[leftmargin=20px]
    \item First, our experimental results demonstrate that the quality of images generated by the teacher model with attention guidance is inferior to that of the original model when we choose to add Gaussian noise to destroy high-attention regions. We hypothesize that since predictive noise plays a crucial role in determining the final content and structure distribution, uncontrollable random noise interference may result in generated images that deviate from the original in terms of content and structure, ultimately leading to method failure. 
    \item Second, through the use of a Gaussian smoothing filter, the attention values of the maximum activated block are influenced by its neighboring blocks. Essentially, after the Gaussian filter is applied to the denoised prediction, every block of pixels within the image can be considered a linear combination of its neighboring pixels. As a result, it efficiently prevents incorrect high-attention guidance that may occur at certain time points.
    \item Third, through further experiments, we observed that, in most cases during sampling, the blurred denoised prediction $\textrm{Gaussian\text-Blur}(\bm{\hat{x}}_0^{t})$ is closer to the final generated sample compared to the original denoised prediction $\bm{\hat x}_0^t$ (Figure~\ref{blur_x0}). Moreover, we attempt to apply Gaussian blur after performing DDIM inverse, but the improvement is worse than directly blurring the denoise prediction $\bm{\hat x}_0^{t}$.
\end{itemize}

\subsection{Predicted Noise Discussion}
A well-trained diffusion model should ensure that the predicted noise $\bm \epsilon_\theta$ obtained from $\bm x_t$ at different $\bm t$ is as consistent as possible when $\bm x_0$ and the noise $\bm \epsilon$ are fixed. 
However, there exists a notable gap between the predicted noise $\bm \epsilon_\theta$ obtained from varying $\bm x_0$ and holding $\bm \epsilon$ and $\bm t$ constant. Specially, diffusion features of $\epsilon_\theta(\bm {x}_t, t)$ from decoder in pre-trained diffusion model will correspond to the subject of $\bm x_0$ at different $t$, as shown in Figure~\ref{delta_eps_l1}. This phenomenon suggests the possibility that pre-trained diffusion models could serve as zero-shot classifiers. We will investigate in future work.

\subsection{Limitation}
\begin{itemize}[leftmargin=20px]
\item 
Due to the unsupervised nature of attention guidance, detecting and correcting instances of incorrect self-attention direction promptly is a challenging task.
Despite efforts to resolve this issue, it remains prevalent, particularly on small datasets. The effective resolution of this issue may require further research on the internal implementation of the diffusion model network.

\item Although the student with semantic gradient predictor achieves better performance and training efficiency, the inference time is higher than the original student model due to the extra predictor. 

\item Since this distillation framework only focuses on scattered time points, it cannot capture the full path of the sample generation. Therefore, it may not be suitable for tasks that depend on path properties (e.g., image-controlled editing). 
\end{itemize}

\subsection{Potential Negative Societal Impacts}

The potential negative impacts of our proposed work are mainly related to deepfakes. Criminals may utilize the power-generating capability of large-scale artificial intelligence models to synthesize false information images, videos and other media. These could be used for illegal activities, such as fraud, forgery, and information pollution. Additionally, deepfakes may lead to information leakage and infringement on individuals’ privacy. To prevent these problems, researchers have developed techniques for detecting deepfakes and proposed various solutions such as blockchain, digital signature technologies and encryption technology to verify the sources of media and ensure the credibility and integrity of information.

\section{Related Work}
\lblapp{related-work}
\subsection{Parameters Conversion}
\lblapp{parameters-conversion}
Suppose we make $x_t = \mu_t x_0+\sigma_t \epsilon$, and the weighting function log-SNR $\lambda_t$ is $w(\lambda_t) = \mathrm {exp}(\lambda_t)$, where $\lambda_t = \log (\mu_t/\sigma_t)$. We can consider another weighting function $\phi_t = \mathrm {arctan}(\sigma_t / \mu_t)$. Since the diffusion process is variance preserving (VP), we have $\mu_\phi = \mathrm{cos}(\phi)$ and $\sigma_\phi = \mathrm{sin}(\phi)$, so $x_\phi = \mathrm{cos}(\phi)x_0 + \mathrm{sin}(\phi)\epsilon$. We can define the velocity of $x_\phi$ as~\cite{salimans2022progressive}: 
\begin{equation}
\begin{aligned}
v_\phi \equiv \frac{dx_\phi}{d\phi} = \mathrm {cos}(\phi)\epsilon - \mathrm{sin}(\phi)x_0
\end{aligned}
\end{equation}
Substituting $\epsilon = [x_\phi - \mathrm{cos}(\phi)x_0] / \mathrm{sin}(\phi)$, the collation yields: 
\begin{equation}
\begin{aligned}
x_0 = \mathrm{cos}(\phi)x_\phi - \mathrm{sin}(\phi)v_\phi
\end{aligned}
\end{equation}
Following the same approach, we get:
\begin{equation}
\begin{aligned}
\epsilon = \mathrm{sin}(\phi)x_\phi + \mathrm{cos}(\phi)v_\phi
\end{aligned}
\end{equation}
Let $\phi_s = \phi_t - \delta$, further using trigonometric identity, we can rewrite the updating rule as
\begin{equation}
\begin{aligned}
x_{\phi_s} &= \mathrm{cos}(\phi_s)\hat{x}_\theta(x_{\phi_t}) + \mathrm{sin}(\phi_s)\hat{\epsilon}_\theta(x_{\phi_t})
\\ &= \mathrm{cos}(\phi_s)(\mathrm{cos}(\phi_t)x_{\phi_t} - \mathrm{sin}(\phi_t)\hat{v}_\theta (x_{\phi_t})) 
\\ & \qquad + \mathrm{sin}(\phi_s)(\mathrm{sin}(\phi_t)x_{\phi_t} + \mathrm{cos}(\phi_t)\hat{v}_\theta (x_{\phi_t}))
\\ &= \mathrm{cos}(\phi_s - \phi_t)x_{\phi_t} + \mathrm{sin}(\phi_s - \phi_t)\hat{v}_\theta (x_{\phi_t})
\\ &= 
\mathrm{cos}(\delta)x_{\phi_t} - \mathrm{sin}(\delta)\hat{v}_\theta (x_{\phi_t})
\end{aligned}
\end{equation}

\subsection{DDIM under ODE framework}
DDIM is shown to be a first-order discrete numerical solution~\cite{lu2022dpm} of ODE~\cite{song2020score}: 
\begin{equation}\label{ode-1}
\begin{aligned}
\frac{d\bm x_t}{dt} = f(t)\bm x_t - \frac{1}{2}g^2(t)\nabla_{\bm x}\log p_t(\bm x_t) ,
\bm x_T \sim \mathcal{N}(0,I)
\end{aligned}
\end{equation}
where the marginal distribution of $\bm x_t$ is $q(\bm x_t|\bm x_0)$. Assuming that $x_t = \mu_t x_0+\sigma_t \epsilon$, and $\nabla_{\bm x}\log p_t(\bm x_t) \approx - \frac{\bm \epsilon_\theta(\bm x_t, t)}
{\sigma_t}$ is approximated by a neural network in~\cite{song2020score}. Let $\lambda_t=\log[\mu_t^2/ \sigma_{t}^2]$, then $f(t) = \frac{1}{2}\sigma_t^2 \frac{d \lambda_t}{dt}$, $g^2(t)=-\sigma_t^2\frac{d\lambda_t}{dt}$. The solution of ODE at time $t$ can be further as:
\begin{equation}
\begin{aligned}
\bm x_t = \frac{\mu_t}{\mu_s}\bm x_s -\frac{1}{2}\mu_t \int_s^t\bigg(\frac{d\lambda_\delta}{d\delta}\bigg)\frac{\sigma_\delta}{\mu_\delta}\bm\epsilon_\theta(x_\delta, \delta)d\delta
\end{aligned}
\end{equation}
To approximate the integral term, one approach is to use a high-order Taylor series expansion. When the order $k=1$, the first term (A) of Eq.(\ref{ode-4}) is just equivalent to the inverse sampling expression of DDIM discretization. $\bm \epsilon_\theta(\bm x_\delta, \delta)$ is assumed to be constant from s to t.
\begin{equation}\label{ode-4}
\begin{aligned}
\bm x_t & = \frac{\mu_t}{\mu_s}\bm x_s -\frac{1}{2}\mu_t\bm\epsilon_\theta(x_\delta, \delta)\int_s^t\bigg(\frac{d\lambda_\delta}{d\delta}\bigg)\frac{\sigma_\delta}{\mu_\delta}d\delta+\mathcal{O}\big((\lambda_t-\lambda_s)^2\big) \\
& = \underbrace{\frac{\mu_t}{\mu_s}\bm x_s -\frac{1}{2}\mu_t(e^{\lambda_t-\lambda_s}-1)\bm\epsilon_\theta(x_\delta, \delta)}_{A}+\mathcal{O}\big((\lambda_t-\lambda_s)^2\big)
\end{aligned}
\end{equation}

\subsection{Conditional Diffusion}
The class-conditional diffusion model~\cite{nichol2022glide, ramesh2022hierarchical} has a wide range of uses in several fields. Dhariwal et al.~\cite{dhariwal2021diffusion} proposes to bootstrap the noise obtained from DDIM sampling predictions by using the gradient $\nabla_{\bm{x}_t} \log p_{\phi}(\bm{c} | \bm{x}_{t})$ provided by an additional pre-trained classifier $p_{\phi}(\bm{c} | \bm{x}_{t})$: 
\begin{equation}\label{classifier_guided_eps}
    \begin{aligned}
        \hat{\bm{\epsilon}}_{\theta}(\bm{x}_{t},t) = \bm{\epsilon}_{\theta}(\bm{x}_{t},t) - \sigma_t \cdot \nabla_{\bm{x}_{t}} \log p_{\phi}(\bm{c} | \bm{x}_{t}) 
    \end{aligned}
\end{equation}
Ho et al.~\cite{ho2022classifier} deformed the expression based on that work and used hard-earned labels $\bm c$ instead of image classifier to obtain the following classifier-free guidance equation: 
\begin{equation}\label{classifier_free_guided_eps}
    \begin{aligned}
    \hat{\bm \epsilon}_{{\bm\theta}}^{\omega}=(1+\omega)\hat{\bm \epsilon}_{c, {\bm\theta}}-\omega\hat{\bm \epsilon}_{{\bm\theta}}
    \end{aligned}
\end{equation}
where $\omega$ is the weight parameter used to control the guidance strength, and $c$ is not restricted to a single kind of information. 
Additionally, Zhang et al.(2023) proposed a novel and flexible conditional diffusion model called ShiftDDPM. By utilizing additional latent space and assigning a diffusion path to each condition based on specific transfer rules, ShiftDDPM disperses the modeling of conditions across all time periods, thereby enhancing the learning capability of the model.

\subsection{Progressive Distillation Model}
\lblapp{progressive-distillation-model}

The Progressive Distillation model (PD) halves the required sampling steps in each iteration by distilling the slower-sampling teacher model into the faster-sampling student model. In each iteration, one DDIM step of the student is made to match two steps of the teacher.

We still use $x_t = \mu_t x_0+\sigma_t \epsilon$, and let $t^{'} = t - 0.5/N$ and $t^{''} = t - 1/N$, where $N$ is the sampling step of the student model and $t$ is the sampling time. Next, we instruct the teacher to perform a two-step DDIM:
\begin{equation}
\begin{aligned}
x_{t^{'}} = \mu_{t^{'}}\hat{x}_\eta(x_t) + \frac{\mu_{t^{'}}}{\sigma_{t}}(x_t - \mu_t \hat{x}_\eta(x_t))
\end{aligned}
\end{equation}
\begin{equation}
\begin{aligned}
x_{t^{''}} = \mu_{t^{''}}\hat{x}_\eta(x_{t^{'}}) + \frac{\sigma_{t^{''}}}{\sigma_{t^{'}}}(x_{t^{'}} - \mu_{t^{'}} \hat{x}_\eta(x_{t^{'}}))
\end{aligned}
\end{equation}
The predicted target value of the teacher model can be set as 
\begin{equation}
\begin{aligned}
\tilde{x}_0^{\mathrm{target}} = \frac{x_{t^{''}} - (\sigma_{t^{''}} / \sigma_{t})x_t}{\mu_{t^{''}} - (\sigma_{t^{''}} / \sigma_{t})\mu_t}
\end{aligned}
\end{equation}
Subsequently, the student model is adjusted to match $\tilde{x}_0^{\mathrm{target}}$, followed by the optimization of the loss function.
\begin{equation}
\begin{aligned}
\mathcal{L}_\theta & = w(\lambda_t)||\tilde{x}_0^{\mathrm{target}} - \hat{x}_0(x_t, t; \theta)||^2_2 \\
&= \max\big(1, \frac{\mu_t^2}{\sigma_t^2}\big)||\tilde{x}_0^{\mathrm{target}} - \hat{x}_0(x_t, t; \theta)||^2_2 
\end{aligned}
\end{equation}
Finally, the fitted student model is used as the new teacher model, and the sampling step $N$ is halved for the next round of iterations. It should be noted that this distillation method is not suitable for use with random sampling in DDPM. The method improves the student model's ability to match by sacrificing generality.

\subsection{Consistency Model}
\lblapp{consistency-model}


Since ODE has been established, there is a unique $x_0$ for any $(x_t, t)$ pair on the same trajectory mapping. The consistency model attempts to directly fit this mapping by defining the function $f(x_t, t) = x_\zeta$, where $\zeta$ represents a small time close to zero and is not set to 0 to avoid special cases at boundary moments.

To ensure the identity of $f(x_\zeta,t) = x_\zeta$, we define it as $f_\theta(x_t, t) = c_{skip}(t)x_t + c_{out}(t)F_\theta(x_t, t)$, where $c_{skip}$ and $c_{out}$ are differentiable functions, such as $c_{skip}(\zeta) = 1, c_{out}(\zeta) = 0$. The trained model enables the generation of high-quality images from random noise through a single NFE using $f_\theta(x_t, t)$. In our experiment, we primarily refer to a consistency model that implements knowledge distillation using pre-trained diffusion models (Consistency Distillation, CD).

Regarding the training process, CD divides the time horizon into N-1 sub-intervals, with boundaries $t_1 = \zeta < t_2 < \cdots < t_{N}=T$. $x_{t_{n+1}}$ is the image obtained from the forward diffusion of the real data $x$. Combined with the ODE solver we could calculate $\hat{x}_{t_n}^\phi$.
\begin{equation}
    \begin{aligned}
    \label{eq:odesolve}
    \hat{x}_{t_n}^\phi = & x_{t_{n+1}} + (t_n - t_{n+1})\Phi(x_{t_{n+1}}, t_{n+1}; \phi) \\ & \qquad x_{t_{n+1}} \sim \mathcal{N}(x; t_{n+1}^2 \mathbf I)
    \end{aligned}
\end{equation}
The function $\Phi(x_{t_{n+1}}, t_{n+1}; \phi)$ denotes the update function (the teacher model). For example, the Euler solver could be represented by $\Phi(x, t; \phi) = -t s_\phi(x, t)$. The objective of the training is to minimize the differences in output $f$ among randomly generated data points obtained on the same path. Further incorporating exponential moving average (EMA) parameters $\theta^{-}$ in the training process to update the target network could enhance the training network's stability and reduce convergence time. $f_{\theta^-}(\hat{x}_{t_n}^\phi, t_n)$ can be regarded as $\tilde{x}_0^{\mathrm{target}}$.When the model training converges, $\theta = \theta^{-}$.
\begin{equation}
\begin{aligned}
\mathcal{L}_\theta = & arg\,\min 
\\ &
\mathbb{E}[\omega(\lambda_t)||f_\theta(x_{t_{n+1}}, t_{n+1}), f_{\theta^-}(\hat{x}_{t_n}^\phi, t_n)||_{\ell_1,\ell_2}]
\end{aligned}
\end{equation}
\begin{equation}
\begin{aligned}
\theta^{-} \leftarrow \mathrm{stopgrad}(\rho \theta^{-} + (1-\rho)\theta)
\end{aligned}
\end{equation}

\section{Additional Results and Samples}
\lblapp{additional-samples}

\begin{figure*}[H]
    \centering
    \includegraphics[width = .9\textwidth]{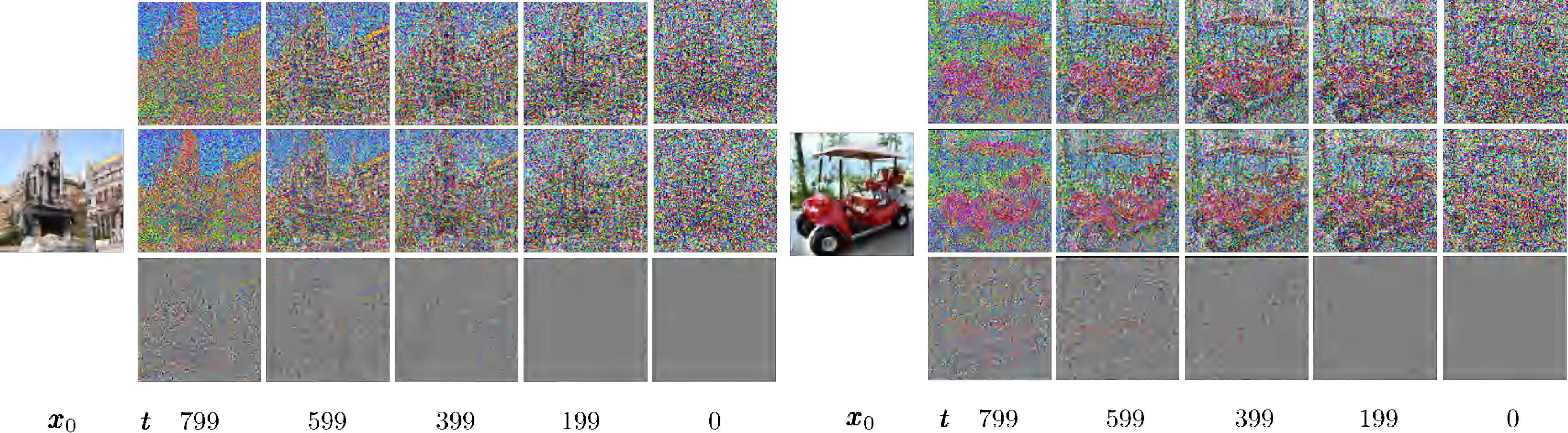}
    \caption{The noise visualization using pre-trained $\epsilon$-prediction diffusion model on ImageNet 128$\times$128. The first row is $\bm \epsilon_eta$. The second row is $\bm {\tilde \epsilon}_{\eta}^{\mathrm {attn}}$. The third row is $\|\bm {\tilde \epsilon}_{\eta}^{\mathrm {attn}} - \bm \epsilon_\eta \|$}
    \label{comparison_eps}
\end{figure*}

\begin{figure*}[H]
    \centering
    \includegraphics[width = .9\textwidth]{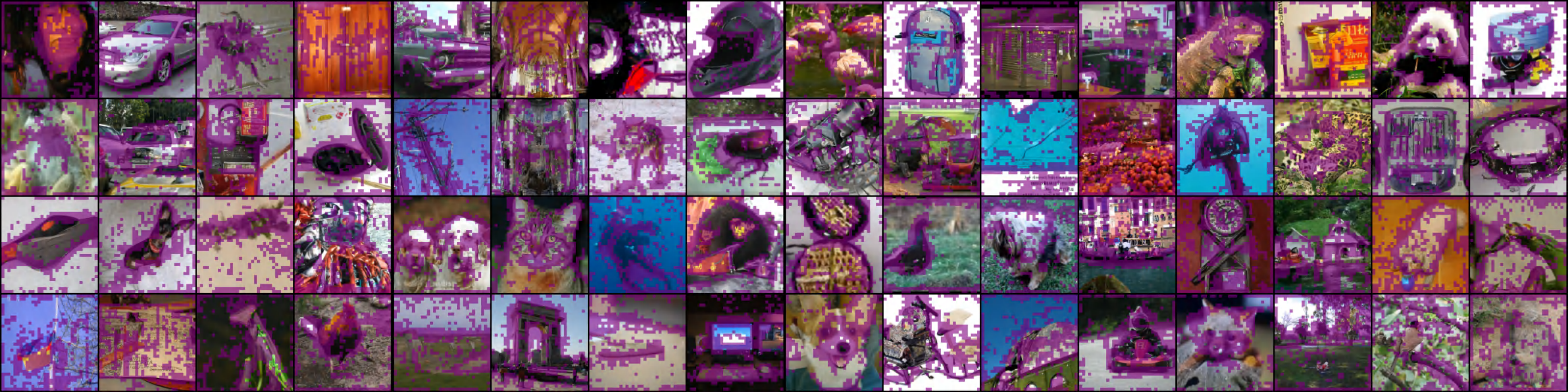}
    \caption{The attention score threshold masking images on ImageNet 128$\times$128. It can be observed that regions with high attention scores tend to concentrate on the edges of the main objects within an image.}
    \label{attn_mask}
\end{figure*}

\begin{figure*}[H]
    \centering
    \includegraphics[width = .8\textwidth]{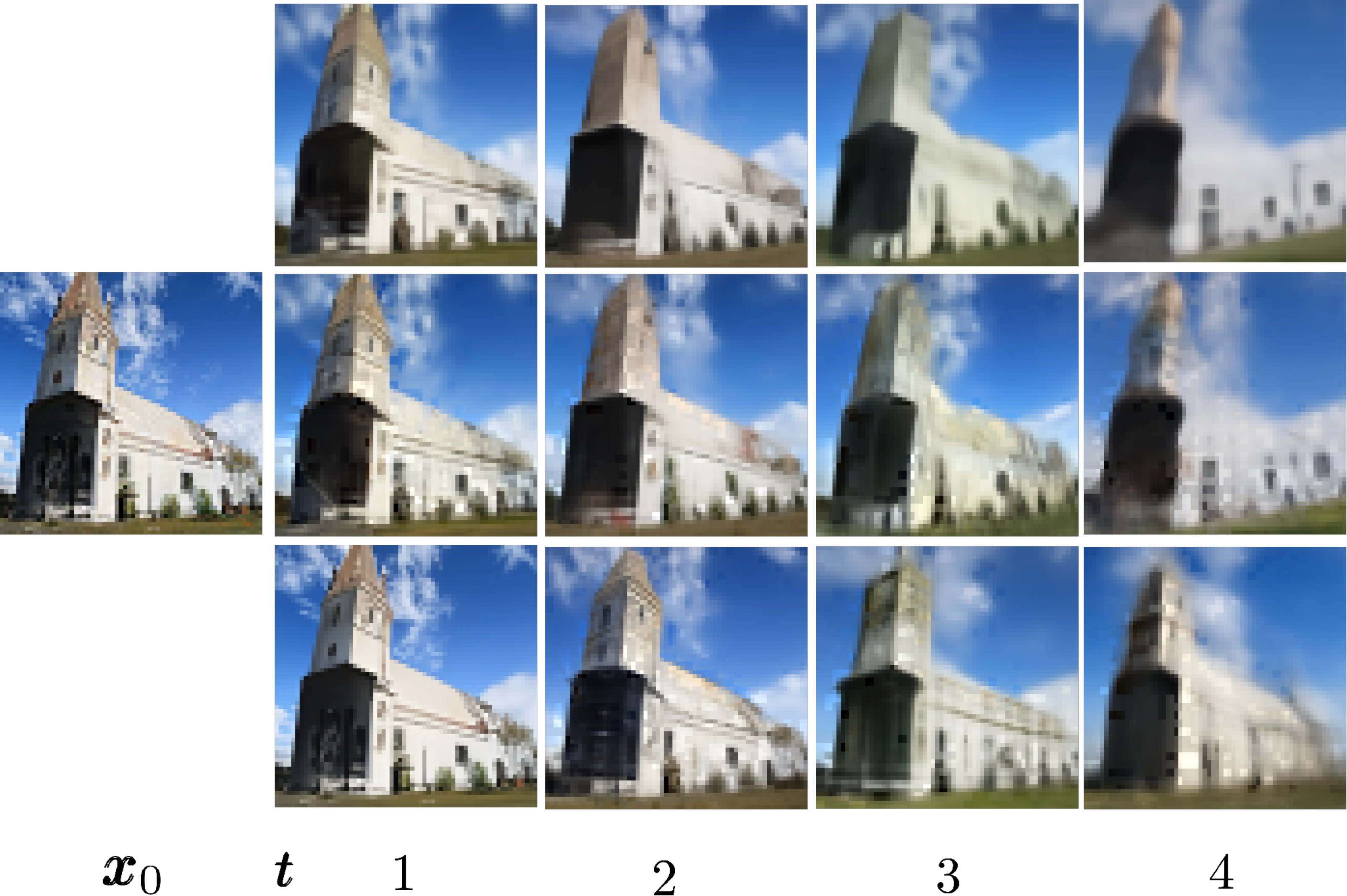}
    \caption{The images of denoised prediction $\bm \hat{x}_0^t$ by denoising $\bm x_t$ at time $t$ on ImageNet 64$\times$64. The first-row use CD, the second-row use CD with attention guidance method and the third-row use SFERD-CD.}
    \label{app_x0}
\end{figure*}

\end{document}